%% file: eMC.tex
\documentclass[twoside,11pt]{article}
\usepackage{jmlr2e}
\usepackage{times}
\usepackage{amsmath,amsfonts,amscd,amsthm,amssymb}
\usepackage{graphicx,subfigure,epstopdf,float} 
\usepackage{natbib}
\usepackage{algorithm}
\usepackage{algorithmic}
\usepackage[symbol]{footmisc}

\usepackage{array,booktabs,paralist}
\newtheorem{theorem}{Theorem}
\newtheorem{lemma}{Lemma}
\newtheorem{assumption}{Assumption}
\newtheorem{definition}{Definition}
\newtheorem{corollary}{Corollary}

\newcommand{\Nu}{\mathcal{V}}
\newcommand{\mybox}{\hfill\(\Box\)}
\newcommand{\kappamin}{\mu_{\mathcal{L}}}
\newcommand{\PsiM}{\Psi(\mathcal{\overline{M}})}
\newcommand{\PsiMsq}{\Psi^2(\mathcal{\overline{M}})}
\begin{document} 
\setlength{\abovedisplayskip}{1pt}\setlength{\belowdisplayskip}{1pt}
\sloppy

\title{Exponential Family Matrix Completion under Structural Constraints}
\author{\name Suriya Gunasekar \email suriya@utexas.edu \AND \name Pradeep Ravikumar \email pradeepr@cs.utexas.edu \AND \name Joydeep Ghosh \email ghosh@ece.utexas.edu}

\maketitle

\begin{abstract} 
\input{abstract}
\end{abstract} 
\input{intro}%\clearpage
\input{setup}

\input{mainResult}
\input{proof}
\input{experiments}
\acks{The research was funded by NSF Grants IIS-0713142 and IIS-1017614. Pradeep Ravikumar acknowledges the support of ARO via W911NF-12-1-0390 and NSF via IIS-1149803, IIS-1320894, DMS-1264033.}
\clearpage

\bibliography{bibliography,latestbib}
\include{appendix}
\end{document}

%% file: abstract.tex
We consider the matrix completion problem of recovering a structured matrix from noisy and partial measurements. Recent works have proposed tractable estimators with strong statistical guarantees for the case where the underlying matrix is low--rank, and the measurements consist of a subset, either of the exact individual entries,  or of the entries perturbed by additive Gaussian noise, which is thus implicitly suited for thin--tailed continuous data. Arguably, common applications of matrix completion require estimators for (a) heterogeneous data--types, such as skewed--continuous, count, binary, etc., (b) for heterogeneous noise models (beyond Gaussian), which capture varied uncertainty in the measurements, and (c) heterogeneous structural constraints beyond low--rank, such as block--sparsity, or a superposition structure of low--rank plus elementwise sparseness, among others. In this paper, we provide a vastly unified framework for generalized matrix completion by considering a  matrix completion setting wherein the matrix entries are sampled from any member of the rich family of \textit{exponential family distributions}; and impose general structural constraints on the underlying matrix, as captured by a decomposable norm regularizer $\mathcal{R}(.)$. We propose a simple convex regularized $M$--estimator for this generalized framework, and provide a unified and novel statistical analysis for this  class of estimators. We finally corroborate our theoretical results on simulated datasets. 

%% file: intro.tex
\section{Introduction}\label{sec:intro} 
In the general problem of matrix completion, we seek to recover a structured matrix from noisy and partial measurements. This problem class encompasses a wide range of practically important applications such as recommendation systems, recovering gene--protein interactions, and analyzing document collections in language processing, among others. In recent years, leveraging developments in sparse estimation and compressed sensing, there has been a surge of work on computationally tractable estimators with strong statistical guarantees, specifically for the setting where a subset of entries of a low--rank matrix are observed either  deterministically, or perturbed by additive noise that is Gaussian \cite{candes2010matrix}, or more generally sub--Gaussian \cite{keshavan2010noise,negahban2012restricted}. While such a Gaussian noise model is amenable to the subtle statistical analyses required for the ill--posed problem of matrix completion, it is not always practically suitable for all data settings encountered in matrix completion problems. For instance, such a Gaussian error model might not be appropriate in recommender systems based on movie ratings that are either binary (likes or dislikes), or range over the integers one through five. The \textit{first question} we ask in this paper is whether we can generalize the statistical estimators for matrix completion  as well as their analyses to general noise models? Note that a noise model captures the uncertainty underlying the matrix measurements, and is thus an important component of the problem specification given any application; and it is thus vital for broad applicability of the class of matrix completion estimators to extend to general noise models. 

Though this might seem like a narrow technical, although important question, it is related to a broader issue. A Gaussian observation model implicitly assumes the matrix values are continuous--valued (and that they are thin--tail--distributed). But in modern applications, matrix data span the gamut of heterogeneous data--types, for instance, skewed--continuous, categorical--discrete including binary, count--valued, among others. This, thus gives rise to the \textit{second question} of whether we can generalize the standard matrix completion estimators and statistical analyses, suited for thin--tailed continuous data, to more heterogeneous data--types? Note that there has been some recent work for the specific case of binary data by~\citet{davenport2012bit}, but generalizations to other data--types and distributions is largely unexplored. 

The recent line of work on matrix completion, moreover, enforces the constraint that the underlying matrix be either exactly or approximately low--rank. Aside from the low--rank constraints, further assumptions to eliminate overly ``spiky" matrices are required for well--posed recovery under partial measurements~\cite{candes2009exact}. Early work provided generalization error bounds  for various low--rank matrix completion algorithms, including algorithms based on nuclear norm minimization~\cite{candes2009exact,candes2010power,
candes2010matrix,recht2011simpler}, max--margin matrix factorization~\cite{srebro2004maximum}, spectral algorithms~\cite{keshavan2010matrix,keshavan2010noise}, and alternating minimization~\cite{jain2013low}. These work made stringent matrix incoherence assumptions to avoid ``spiky" matrices. These assumptions have been made less stringent in more recent results~\cite{negahban2012restricted}, which moreover extend the guarantees to approximately low--rank matrices.  Such (approximate) low--rank structure is one instance of general structural constraints which are now understood to be necessary for consistent statistical estimation under high--dimensional settings (with very large number of parameters and very few observations). Note that the high--dimensional matrix completion problem is particularly ill--posed, since the measurements are typically both very local (e.g. individual matrix entries), and partial (e.g. covering a decaying fraction of entries of the entire matrix). However, the specific (approximately) low--rank structural constraint  imposed in the past work on matrix completion  %, might not always be a realistic assumption for the wide array of applications of matrix completion however. Moreover, it 
does not  capture the rich variety of other qualitatively different structural constraints such as row--sparseness, column--sparseness, or a superposition structure of low--rank plus elementwise sparseness, among others. For instance, in the classical introductory survey on matrix completion~\cite{laurent2009matrix}, the authors discuss structural constraints of a \emph{contraction matrix}, and a \emph{Euclidean distance matrix}.
Thus, the \textit{third question} we ask in this paper is whether we can generalize the recent line of work on low--rank  matrix completion to the more general structurally constrained case.

In this paper, we answer all of the three questions above in the affirmative, and provide a vastly unified framework for generalized matrix completion. We address the first two questions by considering a general matrix completion setting wherein observed matrix entries are sampled from any member of a rich family of \textit{natural exponential family distributions}. Note that this family of distributions encompass a wide variety of popular distributions including Gaussian, Poisson, binomial, negative--binomial, Bernoulli, etc.
Moreover, the choice of the exponential family distribution can be made depending on the form of the data. For instance, thin--tailed continuous data is typically modeled using the Gaussian distribution; count--data is modeled through an appropriate distribution over integers (Poisson, binomial, etc.), binary data through Bernoulli, categorical--discrete through multinomial, etc. We address the last question by considering general structural constraints upon the underlying matrix, as captured by a general regularization function $\mathcal{R}(.)$. Our general matrix completion setting thus captures heterogeneous noise--channels, for heterogeneous data--types, and heterogeneous structural constraints.

In a key contribution, we propose a simple regularized convex $M$--estimator for recovering the structurally constrained underlying matrix in this general setting; and moreover provide a unified and novel statistical analysis for our general matrix completion problem. Following a standard approach~\cite{negahban2012structured}, we (a) first showed that the negative log--likelihood of the subset of observed entries satisfies a form of Restricted Strong Convexity (RSC)~(Definition~\ref{def:rsc}); and (b) under this RSC condition, our proposed $M$--estimator satisfies strong statistical guarantees. We note that proving these individual components for our general matrix completion problem under general structural constraints required a fairly delicate and novel analysis, particularly the first component of showing the RSC condition, which we believe would be of independent interest. A key corollary of our general framework is matrix completion under sub--Gaussian samples and low--rank constraints, where we show that our theorem recovers results comparable to the existing literature~\cite{candes2010matrix,keshavan2010noise,negahban2012restricted}. Finally, we corroborate our theoretical findings via simulated experiments.

%A drawback of our algorithm over alternating minimization algorithms for  matrix completion is that optimizing the proposed convex program involves a nuclear norm constraint, which leads to higher computation time compared to alternating minimization. However, faster algorithms for nuclear norm minimization has been recently developed, which can substantially speed up the computation \cite{Liu2009,Toh2010}. 

%The task of matrix completion from partially observed entries, is in general an ill--posed, however, under structural constraints, guarantees for accurate recovery have been shown in the past for some special cases of our unified framework, specifically under the low--rank assumption on parameter matrix with Gaussian~\cite{care09,capl09,kemo10,kemo10a} or Bernoulli~\cite{davenport2012} assumptions on the observed samples. 

%We generalizes the existing literature in two ways: 
%First, we extend the Gaussian or Bernoulli assumptions on the observed entries %with a broader family of expoential family distributions. 
%Secondly, we provide results for the case when the
%Page 1
%motivate problem

%contributions

\subsection{Notations and Preliminaries}
In this subsection we describe the notations and definitions frequently used throughout the paper. 
%Vectors are denoted by boldface letter, $\mathbf{u}$, $\mathbf{v}$, etc. The $i^{th}$ element of a vector $\mathbf{u}$ is denoted by $\mathbf{u_i}$. 
Matrices are denoted by capital letters, $X$, $\Theta$, $M$, etc. For a matrix $M$, $M_j$ and $M^{(i)}$ are the $j^{th}$ column and $i^{th}$ row of $M$ respectively, and $M_{ij}$ denotes the $(i,j)^{th}$ entry of $M$. %The \emph{transpose} of a matrix, $M$, is denoted by $M^\dagger$. The \textit{trace} and \textit{rank} of a matrix $M$ are denoted by $\text{tr}(M)$ and $\text{rk}(M)$, respectively. The inner product between two matrices is given by $\langle X,Y\rangle=\text{tr}(X^\dagger Y)$.
The \emph{transpose}, \textit{trace}, and \textit{rank} of a matrix $M$ are denoted by $M^\dagger$, $\text{tr}(M)$, and $\text{rk}(M)$, respectively. The inner product between two matrices is given by $\langle X,Y\rangle=\text{tr}(X^\dagger Y)=\sum_{(i,j)}X_{ij}Y_{ij}$.
%The \emph{Singular Value Decomposition} of a matrix $M \in \mathbb{R}^{m\times n}$, of rank $k$ is given my $M=U\Sigma V^\dagger$, where, $U\in\mathbb{R}^{m\times k}$ and $V\in\mathbb{R}^{n\times k}$ are the left and right singular matrices which have orthonormal columns, $U^\dagger U=I_m$, $V^\dagger V=I_n$, and $\Sigma=diag(\sigma_1,\sigma_2,\ldots,\sigma_k)$ is the matrix of singular values. 

For a matrix $M\in\mathbb{R}^{m\times n}$ of rank $r$, with singular values  $\sigma_1\ge\sigma_2\ge\ldots\sigma_r$, commonly used matrix norms include the \textit{nuclear norm} $\|M\|_*=\sum_{i}\sigma_i$, the \textit{spectral norm} $\|M\|_2=\sigma_1$, the \textit{Frobenius norm} $\|M\|_F=\sqrt{\sum_i\sigma_i^2}$, and the \textit{maximum norm} $\|M\|_{\max}={\max_{(i,j)}}\;M_{ij}$. 

Given any matrix norm $\|\cdot\|$, the \textit{dual norm}, $\|\cdot\|^{*}$ 
is given by $\|X\|^{*}={\text{sup}}_{\|Y\|\le 1}\langle X,Y\rangle$.

%is given by:
%\[\|X\|^{*}=\underset{\|Y\|\le 1}{\text{sup}}\;\langle X,Y\rangle
%\]

\begin{definition}[Natural Exponential Family]
\normalfont A distribution of a random variable $X$, in a normed vector space is said to belong to the \emph{natural exponential family}, if its probability density function, characterized by the parameter $\Theta$ in the dual vector space, is given by:	
\[ 
    P(X|\Theta) = h(X)\exp\Big(\langle X,\Theta\rangle -G(\Theta)\Big), 
\]  
where $G(\Theta)=\log\int_Xh(X)e^{\langle X,\Theta\rangle}\text{dX}$, called the log--partition function, is strictly convex, and analytic.

\label{def:expDist}
\end{definition}

\begin{definition}[Bregman Divergence]
\normalfont Let $\phi:\text{dom}(\phi)\to \mathbb{R}$ be a strictly
convex function differentiable in the relative interior of $\text{dom}(\phi)$. The \textit{Bregman divergence} (associated with $\phi$) between $x\in\text{dom}(\phi)$ and $y\in\text{ri}(\text{dom}(\phi))$ is defined as:\[B_\phi(x,y)=\phi(x)-\phi(y)-\langle \nabla \phi(y),x-y\rangle.\]

\label{def:bd}
\end{definition}
\begin{definition} 
[Subspace compatibility constants] 
\normalfont Given a matrix norm $\mathcal{R}(.)$, we define the following maximum and minimum \textit{subspace compatibility} constants of $\mathcal{R}(.)$ w.r.t the subspace $\mathcal{M}$: 
 \[\begin{split}&\Psi(\mathcal{{M}};{\mathcal{R}})
 =\underset{\Theta\in\mathcal{{M}}\setminus\{0\}}{\text{sup}}\frac{\mathcal{R}(\Theta)}{\|\Theta\|_F},\quad
 \Psi_{\min}({\mathcal{R}})
 =\underset{\Theta\neq \{0\}}{\text{inf}}\frac{\mathcal{R}(\Theta)}{\|\Theta\|_F}\end{split}.\]
 Thus, $\forall\Theta\in\mathcal{M}$
\[\Psi_{\min}(\mathcal{R})\|\Theta\|_F\le\mathcal{R}(\Theta)\le\Psi(\mathcal{M},\mathcal{R})\|\Theta\|_F.\]
In the rest of the paper, to avoid notational clutter, the dependence of $\Psi$ and $\Psi_\text{min}$ on $\mathcal{R}$ is suppressed. Thus, $\Psi(\mathcal{{M}};{\mathcal{R}})$ and $\Psi_{\min}({\mathcal{R}})$ are denoted as $\Psi(\mathcal{{M}})$ and $\Psi_{\min}$, respectively.

\label{def:psi}
\end{definition}
\begin{definition} [Restricted Strong Convexity]
\normalfont A loss function $\mathcal{L}$ is said to satisfy \textit{Restricted Strong Convexity} with respected to a subspace $S$, if for some $\kappamin>0$,
\begin{equation*}
\mathcal{L}(\Theta+\Delta)-\mathcal{L}(\Theta)-\langle\nabla
\mathcal{L}(\Theta),\Delta\rangle\ge\kappamin\|\Delta\|_F^2,\forall\Delta\in S.
\end{equation*}
\label{def:rsc}
\end{definition}
%For a linear subspace, $T$, the space orthogonal to $T$ is denoted by $T^\perp$ and the projection of a matrix, $M$ onto $T$ is denoted by $M_T$. We also overload the notation to denote a projection onto a subset, $\Omega$ of indices of the matrix, as $\mathcal{P}_\Omega(M)_{ij}=\left\{\begin{array}{ll} M_{ij} &\text{ if }(i,j)\in\Omega \\ 0 &\text{ otherwise } \end{array}                                                                                                                                                                                                                                                                                             \right.$
\begin{definition} [Sub--Gaussian Distributions]
\normalfont A mean zero random variable $X$ is said to have a sub--Gaussian distribution with parameter $b$ if $\forall\;s>0$, the distribution satisfies $E[e^{sX}]\le e^{s^2b^2/2}$. Further, if $X$ is sub--Gaussian with parameter $b$ and $E[X]=0$, then $\text{Var}(X)\le b^2$ (\citet{vershynin2010introduction}).
\label{def:subGauss}
\end{definition}

%% file: setup.tex
\section{Exponential Family Matrix Completion}\label{sec:eMC}
Denote the underlying target matrix by $\Theta^*\in\mathbb{R}^{m\times n}$. We then assume that individual entries $\Theta^*_{ij}$ are observed indirectly via a noisy channel: specifically, via a sample drawn from the corresponding member of \textit{natural exponential family} ~(see Definition~\ref{def:expDist}):
\begin{align}
P(X_{ij}|\Theta^*_{ij}) &= h(X_{ij}) \, \exp\left(X_{ij}\Theta^*_{ij} -G(\Theta^*_{ij})\right),
\label{eq:expfamind}
\end{align}
where $G:\mathbb{R}\to\mathbb{R}$ is a strictly convex, and analytic function called the log--partition function. 

Consider the random matrix $X\in\mathbb{R}^{m\times n}$, where each entry $X_{ij}$ is drawn independently from the corresponding distribution in \eqref{eq:expfamind}; it can be seen that:
\begin{flalign}
\nonumber P(X|\Theta^*) &=\prod_{ij} \left( h(X_{ij})\, \exp\left(X_{ij}\Theta^*_{ij}-G(\Theta^*_{ij})\right)\right)&\\
&=h(X)\exp\left(\langle X,\Theta^* \rangle - G(\Theta^*)\right),&
\label{eq:expfam}
\end{flalign}
where we overload the notation to denote $G:\mathbb{R}^{m\times n}\to\mathbb{R}$ as $G(\Theta)=\sum_{ij}G(\Theta_{ij})$, and the base measure $h(X)$ as $h(X)=\prod_{ij}{h(x_{ij})}$.
%It can be verified that $G(\Theta)=\log{\left\{\int_Xh(X)e^{\langle X,\Theta\rangle}\text{d}X\right\}}$. 

\textbf{Uniformly Sampled Observations:} 
In a ``fully observed'' setting, we would observe all the entries of the observation matrix $X\in\mathbb{R}^{m\times n}$. However, we consider a partially observed setting, where we observe entries over a subset of indices $\Omega\subset [m]\times[n]$. We assume a uniform sampling model, so that 
\begin{equation}
	\forall \;(i,j)\in\Omega,\;i\sim\text{uniform}([m]),\;j\sim\text{uniform}([n]).
\end{equation}
 Note that, under the above described sampling scheme, an index $(i,j)$ can be sampled multiple times, in such cases we include the multiple instances of $(i,j)$ in $\Omega$ (and not just the unique indices in $\Omega$). Given $\Omega$, we define the following matrix $\mathcal{P}_\Omega(X)$ as $$\mathcal{P}_\Omega(X)=\sum_{(i,j)\in\Omega} X_{ij}e_ie_j^\dagger.$$ 
%Given, $\Omega$, we define the following matrix $\mathcal{P}_\Omega(X)$:\\ $\mathcal{P}_\Omega(X)_{ij}=\left\{\begin{array}{ll}X_{ij}&\text{if }(i,j)\in\Omega\\0&\text{otherwise.}\end{array}\right.$
%\item Assume that $\forall\;(i,j)\in[m]\times[n]$, the matrix element, $X_{ij}$, of $X$, is observed with probability $p$
%\item $\delta_{ij}$ be an indicator that $X_{ij}$ was observed and $\Omega=\{(i,j):\delta_{ij}=1\}$

The matrix completion task can then be stated as the estimation of $\Theta^*$ from $(\Omega, \mathcal{P}_\Omega(X))$, where $X\sim P(X|\Theta^*)$. As noted earlier, this problem is ill--posed in general. However, as we will show, under structural constraints imposed on the parameter matrix $\Theta^*$, we are able to design an $M$--estimator with a near optimal deviation from $\Theta^*$. 

\subsection{Applications}
%Some examples of common members of $1D$ exponential family distributions are described below:
\textbf{Gaussian (fixed  $\sigma^2$)} is typically used to model continuous data, $x\in\mathbb{R}$, such as measurements with additive errors, affinity datasets. Here, $G(\theta)=\frac{1}{2}\sigma^2\theta^2$.\\
\textbf{Bernoulli} is a popular distribution of choice to model binary  data, $x\in\{0,1\}$, with $G(\theta)=\log{(1+e^\theta)}$. Some examples of data suitable for Bernoulli model include social networks, gene protein interactions, etc. \\
\textbf{Binomial (fixed $N$)} is used to model number of successes in $N$ trials. Here, $x\in\{0,1,2,\ldots,N\}$, and  $G(\theta)=N\log{(1+e^\theta)}$. Some applications include predicting success/failure rate, survey outcomes, etc.\\
\textbf{Poisson} is used to model count data $x\in\{0,1,2,\ldots\}$, such as arrival times, events per unit time, click--throughs among others. Here, $G(\theta)= e^\theta$.\\
\textbf{Exponential} is often used to model positive valued continuous data $x\in\mathbb{R}_+$, specially inter arrival times between events. Here, $G(\theta)=-\log{(-\theta)}$. %\begin{table}[H]
%\begin{tabular}{|p{2.6cm}|p{2cm}|>{\raggedright\arraybackslash}p{2.6cm}|}
%\hline
%&$G(\Theta)$& Examples\\ \hline
%Gaussian \break (fixed  $\sigma^2$), $x\in\mathbb{R}$&$\frac{1}{2}\sigma^2\theta^2$& measurements with additive errors, affinity datasets\\
%Bernoulli, \break $x\in\{0,1\}$&$\log{(1+e^\theta)}$&social networks, gene protein interactions\\
%Binomial($N$), $x\in\{0,1,\ldots,N\}$&$N\log{(1+e^\theta)}$&success/failure rate, binary poll results\\
%Poisson, \break $x\in\{0,1,\ldots\}$&$e^\theta$& arrival times, events per unit time, click through\\
%Exponential, \break $x\in\mathbb{R}_+$&$-\log{(-\theta)}$& inter arrival times\\
%\hline
%\end{tabular}
%\label{tab:expfam}
%\end{table}

\subsection{Log--likelihood}
Denote the gradient map:
\[g(\Theta) \triangleq  \nabla G(\Theta) \in\mathbb{R}^{m\times n},\;\text{where}\; g(\Theta)_{ij}=\frac{\partial G(\Theta)}{\partial \theta_{ij}}.\] 
It can then be verified that the mean and variance of the distribution $P(X|\Theta^*)$ are given by $\mathbb{E}[X]=g(\Theta^*)$, and  $\text{Var}(X)=\nabla^2 G(\Theta^*)$, respectively.
The Fenchel conjugate of the log partition function $G$, is denoted by:  $F(X)\triangleq {\text{sup}}_{\Theta}\; \langle X,\Theta\rangle -G(\Theta)$.

A useful consequence of the exponential family is that the negative log--likelihood is convex in the natural parameters $\Theta^*$, and moreover has a bijection with a large class of \textit{Bregman divergences}~(Definition~\ref{def:bd}). The following relationship was first noted by~\cite{forster2002relative}, and later established by~\cite{banerjee2005clustering}~[Theorem~4]:
\begin{equation}
-\log{P(X|\Theta)}\propto B_F(X,g(\Theta)),\;\forall X\in\text{dom(F)}.
\label{eq:bijection}
\end{equation}

\subsection{Discussion and directions for future work}
We consider the standard matrix--completion setting where the distribution of the observation matrix $X$ in \eqref{eq:expfam} corresponds to its entries $X_{ij}$ being drawn independently from the other entries. Further, the probability of observing a specific entry $X_{ij}$, under uniform sampling is independent of the noise channel or the distribution $P(X_{ij}|\Theta^*_{ij})$. 
However, in some applications, it might be beneficial to have a sampling scheme involving dependencies; for instance, when a user gives a movie a bad rating, we might want to vary the sampling scheme to sample an entirely different region of the matrix. It would be interesting to extend the analysis of this paper to such a dependent sampling setting. %We could still consider an exponential family distribution for the observation matrix $X$ but with interaction terms (e.g. pairwise terms) involving multiple entries of $X$.

The form of the observation random matrix distribution in \eqref{eq:expfam}, given the individual observations in \eqref{eq:expfamind}, can be seen to have connotations of multi--task learning: here recovering each individual matrix entry $\Theta^*_{ij}$ together with its corresponding noise model forms a single task, and all these tasks can be performed jointly given the shared structural constraint on $\Theta^*$. It would be interesting to generalize this to form a more general statistical framework for \emph{partial} multi-task learning.

We use the general class of exponential family distributions as the underlying probabilistic model capturing the measurement uncertainties. The richness of the class of exponential family distributions has been used in other settings to provide general statistical frameworks. \citet{kakade2010learning} provide a generalization of compressed sensing problem to general exponential family distributions. Note however that results from compressed sensing cannot be immediately extended to matrix completion case, since the sampling operator $\mathcal{P}_\Omega$ does not satisfy the typical assumptions (restricted isometry or restricted eigenvalues)  made in such settings; see \citep{candes2009exact} for additional discussion. There have been extensions of classical probabilistic PCA~\cite{tipping1999probabilistic} from Gaussian noise models to  exponential family distributions~\citet{collins2001generalization,mohamed2008bayesian,gordon2002generalized}. There have also been recent extensions of probabilistic graphical model classes, beyond Gaussian and Ising models, to  multivariate extensions of exponential family distributions~\citep{yang2012graphical,yang2013conditional}.

More complicated probabilistic models have also been proposed in the context of collaborative filtering~\cite{mnih2007probabilistic,salakhutdinov2008bayesian}, but these typically involve  non--convex optimization, and it is difficult to extend  the rigorous statistical analyses of the form in this paper (and in the matrix completion literature) to these models.

%% file: mainResult.tex
\section{Main Result and Consequences}

As noted in the introduction, we consider the matrix completion setting with general structural constraints on the underlying target matrix $\Theta^*$. To formalize the notion of such structural constraints, we follow \citep{negahban2012structured}, and assume that $\Theta^*$ satisfies $\Theta^*\in\mathcal{M}\subseteq{\overline{\mathcal{M}}}\subset{R}^{m\times n}$, 
for some subspace $\mathcal{M} \subseteq \overline{\mathcal{M}}$, which contains parameter matrices that are  structured similar to the target (the corresponding structural constraints such as low rankness, low rankness+sparsity etc); we also allow the flexibility of working with a superset $\overline{\mathcal{M}}$ of the model subspace that is potentially easier to analyze. Moreover, we use their definition of a decomposable norm regularization function, $\mathcal{R}(.):\mathbb{R}^{m\times n}\to R_+$, which suitably captures these structural constraints:
\begin{assumption}(Decomposable Norm Regularizer)  We assume that $\mathcal{R}(.)$ is a matrix norm, and is decomposable over $({\mathcal{M}},{\overline{\mathcal{M}}}^\perp)$, i.e. if $X\in{\mathcal{M}},\;Y\in{\overline{\mathcal{M}}}^\perp$, then,
\[\mathcal{R}(X+Y)=\mathcal{R}(X)+\mathcal{R}(Y).\]

%\normalfont Numerous examples of such decomposable regularizers are provided in~\cite{negahban2012structured}. 
\normalfont
We provide some examples of such decomposable regularizers and structural constraint subspaces, and refer to \citep{negahban2012structured} for more examples and discussion.

\textbf{Example 1. Low--rank}:  This is a common structure assumed in numerous  matrix estimation problems, particularly those in collaborative filtering, PCA, spectral clustering, etc.
The corresponding structural constraint subspaces $(\mathcal{M},\mathcal{\overline{M}}^\perp)$ in this case correspond to a linear span of specific one--rank matrices; we discuss these in further detail in Section~\ref{sec:mainResult}, where we derive a corollary of our general theorem to the specific case of recovery guarantees for low--rank constrained matrix completion. The nuclear norm  $\mathcal{R}(\Theta)=\|\Theta\|_*=\sum_k\sigma_k$, has been shown to be \emph{decomposable} with respect to these constraint subspaces~\cite{fazel2001rank}.

\textbf{Example 2. Block sparsity}: Another important  structural constraint for a matrix is block--sparsity, where each row is either all zeros or mostly non--zero, and the number of non--zero rows is small. The structural constraint subspaces in this case correspond to a linear span of specific Frobenius--norm--one matrices that are non--zero in a small subset of the rows (dependent on $\Theta^*$); it has been shown that $\ell_1/\ell_q$ $(q>1)$ norms~\cite{negahban2008joint,obozinski2011support} are decomposable with respect to such structural constraint subspaces. Recalling that $\Theta^{(i)}$ is the $i^{th}$ row of $\Theta$, the $\ell_1/\ell_q$ norm is defined as:
\[\|\Theta\|_{1,q}=\sum_{i=1}^m\|\Theta^{(i)}\|_q=\sum_{i=1}^m\Big[\Big(\sum_{j=1}^n |\Theta_{ij}|^q\Big)^{1/q}\Big].
\]

\textbf{Example 3. Low--rank plus sparse}: This structure is often used to model low--rank matrices which are corrupted by a sparse outlier noise matrix. The structural constraint subspaces corresponding to these consist of the linear span of weighted sum of specific rank--one matrices and  sparse matrices with non--zero components on specified positions; and appropriate regularization function decomposable with respect to such structural constraints is the infimum convolution of the weighted nuclear norm with weighted elementwise $\ell_1$ norm, $\|M\|_{1,1}=\sum_{ij}|M_{ij}|$ \cite{candes2011robust,YR13}:
\[\mathcal{R}(\Theta)=\inf\{\lambda_{1} \|S\|_{1,1} + \lambda_{2} \|L\|_*: \Theta=S+L\}.
\]

\label{ass:first}
\end{assumption}
%\begin{assumption} $\mathcal{R}(X)\ge\|X\|_*$, where $\|X\|_*$ is the nuclear norm.
%\normalfont Nuclear norm is the most commonly used convex regularizer to enforce low--rankness. Typically,  structural constraints on matrix data are imposed in addition to the low--rank assumption (like low rank+sparse), in which case this assumption is naturally satisfied.
%\label{ass:greatNuc}
%\end{assumption}
The second assumption we make is on the curvature of the log--partition function. This is required to establish a form of RSC (Definition~\ref{def:rsc}) for the loss function.
\begin{assumption} The second derivative of the log--partition function $G:\mathbb{R}\to\mathbb{R}$ has atmost an exponential decay, i.e, \[\nabla^2G(u)\ge e^{-\eta|u|},\;\forall\;u\in\mathbb{R},\text{ for some  }\eta>0\]

\normalfont It can be verified that commonly used members of natural exponential family obey this assumption.

\label{ass:two}
\end{assumption}

Finally, we make an assumption to avoid ``spiky" target matrices. As \citet{candes2009exact} show with numerous examples, low--rank and presumably other such structural constraints as above, by themselves are not sufficient for accurate recovery, in part due to the infeasibility of recovering overly ``spiky'' matrices with very few large entries. Early work~\cite{candes2010matrix,keshavan2010matrix,keshavan2010noise}, assumed stringent matrix incoherence conditions to preclude such matrices, while more recent work~\cite{davenport2012bit,negahban2012restricted}, relax these assumptions to restricting the \textbf{spikiness ratio}, defined as follows:
\begin{equation}
\alpha_{\text{sp}}(\Theta)=\frac{\sqrt{mn}\|\Theta\|_{\max}}{\|\Theta\|_F}.
\end{equation}

\begin{assumption} There exists a known $\alpha^*>0$, such that 
\[\|\Theta^*\|_{\max}=\frac{\alpha_{sp}(\Theta^*)}{\sqrt{mn}}\|\Theta^*\|_F\le\frac{\alpha^*}{\sqrt{mn}}.\]

\label{ass:last}
\end{assumption}

\subsection{$M$--estimator for Generalized Matrix Completion}\label{sec:alg}
We propose a regularized $M$--estimate as our candidate  parameter matrix $\widehat{\Theta}$. The norm regularizer $\mathcal{R}(.)$ used  is a convex surrogate for the structural constraints, and is assumed to satisfy A~\ref{ass:first}. For a suitable $\lambda>0$,
\begin{align}
\nonumber &&\widehat{\Theta}&=\underset{\|\Theta\|_{\max}\le\frac{\alpha^*}{\sqrt{mn}}}{\text{argmin}} \frac{mn}{|\Omega|}\Big[\sum_{ij\in\Omega}  -\log{P(X_{ij}|\Theta_{ij})} \Big] +\lambda\mathcal{R}(\Theta)\\
&&&=\underset{\|\Theta\|_{\max}\le\frac{\alpha^*}{\sqrt{mn}}}{\text{argmin}} \frac{mn}{|\Omega|}\Big[\sum_{ij\in\Omega}G(\Theta_{ij})-X_{ij}\Theta_{ij}\Big]+\lambda\mathcal{R}(\Theta).
\label{eq:main}
\end{align}
In the above estimator, for simplicity we have assumed that the domain of the minimizing function spans all or $\mathbb{R}^{m\times n}$. In cases where this is violated, additional constraints to restrict $\Theta$ to the domain could be imposed on the estimator and the results and analysis in the following section still hold. 
The above optimization problem is a convex program, and can be solved by any off--the shelf convex solvers. 

\subsection{Main Results} \label{sec:mainResult} Without loss of generality, suppose that $m\le n$. Let $\mathcal{R}^*(.)={\text{sup}}_{\mathcal{R}(X)\le 1}\langle X,.\rangle$ be the dual norm of the regularizer $\mathcal{R}(.)$. Further, let $\PsiM$ and $\Psi_{\min}$ be the maximum and minimum \textit{subspace compatibility constants} of $\mathcal{R}$ w.r.t~the model subspace $\overline{\mathcal{M}}$~(Definition~\ref{def:psi})\footnote{We suppress the dependence of $\Psi$ and $\Psi_{\min}$ on $\mathcal{R}$ in our notation to avoid notational clutter}. We next define the following quantity:
\[\kappa_{\mathcal{R}}(n,|\Omega|) := \mathbb{E}\Big[\frac{\sqrt{mn}}{|\Omega|}\mathcal{R}^*\Big(\sum_{ij\in\Omega}\epsilon_{ij}e_ie_j^*\Big)
\Big],\]
where the expectation is over the random sampling index set $\Omega$, and over the Rademacher sequence $\{\epsilon_{ij}:\forall(i,j)\in\Omega\}$; here $\{e_{i}\in\mathbb{R}^m\}$, $\{e_j\in\mathbb{R}^n\}$ are the standard basis. This quantity $\kappa_{\mathcal{R}}(n,|\Omega|)$ captures the interaction between the sampling scheme and the structural constraint as captured by the regularizer (specifically its dual $\mathcal{R}^*$). Note that it depends only on $n$ ($n\ge m$), and on the size $|\Omega|$ of $\Omega$.

\begin{theorem}
Let $\widehat{\Theta}$ be the estimate from  \eqref{eq:main} with $\frac{\lambda}{2}\ge\frac{mn}{|\Omega|}\mathcal{R}^*(\mathcal{P}_\Omega(X-g(\Theta^*))$. Under the Assumptions~\ref{ass:first}--\ref{ass:last}, if $|\Omega|\ge c_0\PsiMsq n\log{n})$ for large enough $c_0$, then given a constant $\beta>0$, $\exists$ a constant $K_\beta>0$ such that, using $\kappamin:=e^{-\frac{2\eta\alpha^*}{\sqrt{mn}}}\Big(K_\beta-\frac{64}{c_0}\sqrt{\frac{|\Omega|\kappa_{\mathcal{R}}^2(n,|\Omega|)}{n\log{n}}}\Big)$, the following holds with  probability $>1-4e^{-(1+\beta)\Psi^4_{\min}\log^2{n}}$:
%\vspace{1mm}
\small
\[
\|\widehat{\Theta}-\Theta^*\|^2_F\le \Psi^2(\mathcal{\overline{M}})\max\left\{\frac{3\lambda^2}{2\kappamin^2},\frac{c_0^2\alpha^{*2}n\log n}{|\Omega|}\right\},
\]
\normalsize provided $\kappamin>0$.

\label{thm:main}
\end{theorem}
In the above theorem, $\eta$ and $\alpha^*\ge\alpha_{sp}(\Theta^*)\|\Theta^*\|_F$ are constants from Assumptions~\ref{ass:two} and \ref{ass:last}, respectively. 
Our proof uses elements from ~\citet{negahban2012structured}, as well as \citet{negahban2012restricted} where they analyze the case of low--rank structure and additive noise, and  establish a form of restricted strong convexity (RSC) for squared loss over subset of matrix entries (closely relates to the special case, when the exponential family distribution assumed in~\eqref{eq:expfam} is Gaussian). However, showing such an RSC condition for our general loss function over a subset of structured matrix entries involved some delicate arguments. Further, we provide a much simpler proof that moreover only required a low--spikiness condition rather than a multiplicative spikiness and structural constraint. Moreover, we are able to provide an RSC condition broadly for general structures, and the negative log--likelihood loss associated  with general exponential family distribution.

\subsection{Corollary}\label{sec:cor}
An important special case of the problem is when the parameter matrix $\Theta^*$, is assumed to be of a low--rank $r\ll m$. The most commonly used convex regularizer to enforce low--rank is the nuclear norm. The intuition behind the low--rank assumption on the parameter matrix is as follows: the parameter $\Theta_{ij}^*$, can be thought of as the true measure of affinity between the entities corresponding to row $i$ and  column $j$, respectively; and the data $X_{ij}$, is a sample from a noisy channel parametrized by $\Theta_{ij}$. It is hypothesized that $\{\Theta_{ij}^*\}$, are obtained from a weighted interaction of a 
small number of latent variables as, $\Theta_{ij}^*=\sum_{k=1}^r\sigma_ku_{ik}v_{jk}$. 
%Nuclear norm as a convex surrogate for rank constraints was introduced by~\cite{fazel2001rank}, and has been successfully used with provable guarantees for matrix completion under additive noise assumption~\cite{candes2010noise}, and for binary matrix completion~\cite{davenport2012bit}. 

Let $\{\mathbf{u_k}\in\mathbb{R}^m\}$ and $\{\mathbf{v_k}\in\mathbb{R}^n\}$, $k\in[r]$ be the left and right singular vectors, respectively of $\Theta^*$. 
Let the column and row span of $\Theta^*$ be $U^*\triangleq\text{col}(\Theta^*)=\text{span}\{\mathbf{u_i}\}$ and $V^*\triangleq\text{row}(\Theta^*)=\text{span}\{\mathbf{v_j}\}$, respectively.  Define:
\begin{align}\begin{split}
&\mathcal{M}:=\{\Theta:\text{row}(\Theta)\subseteq V^*,\;\text{col}(\Theta)\subseteq U^*\}, \text{ and} \\
&\overline{\mathcal{M}}^\perp:=\{\Theta:\text{row}(\Theta)\subseteq V^{*\perp},\;\text{col}(\Theta)\subseteq U^{*\perp}\}.
\label{eq:M}
\end{split}\end{align}
It can be verified that, $\mathcal{M}\neq\overline{\mathcal{M}}$, however, $\mathcal{M}\subset\overline{\mathcal{M}}$.%The following corollary gives recovery guarantees for special case of low--rank constraints.
\begin{corollary}
Let $\Theta^*$ be a low--rank matrix of rank atmost $r\ll m$. Further,  let $\forall{(i,j)}$, $(X_{ij}-g(\Theta^*_{ij}))$ are sub--Gaussian (Definition~\ref{def:subGauss}) with parameter $b$, and $|\Omega|>c_0rn\log{n}$ for large enough constant $c_0$. Given any $\beta>0$, there exists constants $c_\beta>0$, $C_\beta>0$ and $K_\beta>0$, such that using $\mathcal{R}(.)=\|.\|_*$ and $\frac{\lambda}{2}:=c_\beta\sqrt{mn}b\sqrt{\frac{n\log{n}}{|\Omega|}}$
 in \eqref{eq:main}, w.p.~$>1-4e^{-(1+\beta)\log^2{n}}-e^{-(1+\beta)\log(n)}$,% the following holds for the estimate $\widehat{\Theta}$:
\[
\frac{1}{{mn}}\|\widehat{\Theta}-\Theta^*\|_F^2\le C_\beta\frac{\max\{b^2,\alpha^{*2}/mn\}}{\kappamin^{2}}\left(\frac{rn\log n}{|\Omega|}\right),%\le C_0^{\prime\prime}\frac{\max\{\alpha^{*2},1\}\sigma^2}{\kappamin^{\prime2}},
\]
where  $\kappamin=K_\beta^\prime e^{-\frac{2\eta\alpha^*}{\sqrt{mn}}}>0$.

\label{cor:corollary}
\end{corollary}
\textbf{Remark 1:} Note that the above results hold for the minimizer $\widehat{\Theta}$ of the convex program in \eqref{eq:main}, optimized for any $\alpha^*\ge\alpha_{sp}(\Theta^*)\|\Theta^*\|_F$; in particular it holds with $\alpha^*=\alpha_{sp}(\Theta^*)\|\Theta^*\|_F$, where $1\le\alpha_{sp}(\Theta^*)\le\sqrt{mn}$. While in practice $\alpha^*$ is chosen through cross--validation, the theoretical bound in Corollary~\ref{cor:corollary} can be tightened to the following (if $\|\Theta\|_F\ge1$):
\begin{equation}
\frac{\|\widehat{\Theta}-\Theta^*\|^2_F}{\|\Theta^*\|^2_F}\le C_\beta\frac{\alpha_\text{sp}^2(\Theta^*)\max{\{b^2,1\}}}{\kappamin^{2}}\left(\frac{rn\log n}{|\Omega|}\right).
\end{equation}
Similar bound can be obtained for Theorem~\ref{thm:main}. 

\textbf{Remark 2:} $b^2$  is a measure of noise per entry; $\forall (i,j),\text{Var}(X_{ij}-g(\Theta^*_{ij}))\le b^2$. Note that as we do not make stronger matrix incoherence assumptions, only an approximate recovery is guaranteed even as $b\to 0$.%The sub--Gaussianity assumption in the corollary can be relaxed using other suitable measure of the 

%% file: proof.tex
\section{Proof}\label{sec:proof} 
In this section we provide key steps in the proofs of the main results (Section~\ref{sec:mainResult}-\ref{sec:cor}). Proofs of intermediate lemmata are deferred to the appendix. 

\subsection{Proof of Theorem \ref{thm:main}}
The proof of our main theorem involves two key steps:
\begin{compactitem}
\item We first show that, under assumptions Assumption \ref{ass:first}--\ref{ass:last}, \textit{RSC} of the form in Definition \ref{def:rsc} holds for the loss function in~\eqref{eq:main} over a \emph{large subset} of the solution space.
\item When the RSC condition holds, the result follows from a few simple calculations; we handle the case where RSC does not hold separately.
\end{compactitem}

Let $\widehat{\Delta}=\widehat{\Theta}-\Theta^*$. We state two results of interest.

\begin{lemma} We define the following subset:
\[\Nu=\{\Theta\in\mathbb{R}^{m\times n}:\mathcal{R}(\Theta_{\overline{\mathcal{M}}^\perp})\le 3\mathcal{R}(\Theta_{\overline{\mathcal{M}}})\},\]
where recall $(\mathcal{M},\overline{\mathcal{M}}^\perp)$ from Assumption  \ref{ass:first}, and $\Theta_{\overline{\mathcal{M}}}$ is the projection of $\Theta$ onto the subspace $\mathcal{\overline{M}}$. 
If $\widehat{\Theta}$ is the minimizer of \eqref{eq:main}, and $\frac{\lambda}{2}\ge\frac{mn}{|\Omega|}\mathcal{R}^*(\mathcal{P}_\Omega(X-g(\Theta^*))$, then 
$\widehat{\Delta}=\widehat{\Theta}-\Theta^*\in\Nu$.\\
\normalfont The proof follows from Lemma~$1$ of \citet{negahban2012structured}.\mybox

\label{lem:inNu}
\end{lemma}
\begin{lemma}Let $\widehat{\Theta}$ be the minimizer of \eqref{eq:main}. If $\frac{\lambda}{2}\ge\frac{mn}{|\Omega|}\mathcal{R}^*(\mathcal{P}_\Omega(X-g(\Theta^*))$,  then:
\[ \frac{mn}{|\Omega|}\sum_{(i,j)\in\Omega}B_G(\widehat{\Theta}_{ij},\Theta^*_{ij})\le\frac{3\lambda\Psi(\mathcal{\overline{M}})}{2}\|\Theta^*-\widehat{\Theta}\|_F
\]
\normalfont The proof is provided in Appendix~\ref{app:BFOmega}.\mybox

\label{lem:BFOmega}
\end{lemma}
Recall the notation $\alpha_{\text{sp}}({\Delta})=\frac{\sqrt{mn}\|\Delta\|_{\max}}{\|\Delta\|_F}$. We now consider two cases, depending on whether the following condition holds for the constant $c_0>0$ dictated by Theorem~\ref{thm:main}:
\begin{align} 
	\alpha_{\text{sp}}(\widehat{\Delta}) \le \frac{1}{c_0\PsiM}\sqrt{\frac{|\Omega|}{n\log n}}.
	\label{EqnAlphaCondition}
\end{align}

\noindent \textbf{Case 1:} Suppose condition in \eqref{EqnAlphaCondition} does not hold; so that $\alpha_{\text{sp}}(\widehat{\Delta}) > \frac{1}{c_0\PsiM}\sqrt{\frac{|\Omega|}{n\log n}}$. From the constraints of the optimization problem~\eqref{eq:main}, we have that $\|\widehat{\Delta}\|_{\max}\le\|\widehat{\Theta}\|_{\max}+\|\Theta^*\|_{\max}\le (2\alpha^*/{\sqrt{mn}})$. Thus,
\small
\begin{align}
\begin{split}
\|\widehat{\Delta}\|_F=\frac{\sqrt{mn}\|\widehat{\Delta}\|_{\max}}{\alpha_{\text{sp}}(\widehat{\Delta})}\le2c_0\alpha^*\sqrt{\frac{\Psi^2(\mathcal{\overline{M}}) n\log n}{|\Omega|}}.
\end{split}
\label{eq:case3}
\end{align}
\normalsize

\noindent \textbf{Case 2:} Suppose condition in \eqref{EqnAlphaCondition} does hold. Then, the following theorem shows that an RSC condition of the form in Definition \ref{def:rsc} holds.
%If $\alpha_{\text{sp}}(\widehat{\Delta})\le\frac{1}{c_0\PsiM}\sqrt{\frac{|\Omega|}{n\log n}}$, we show the following:
\begin{theorem} [Restricted Strong Convexity] For $c_0$ given by Theorem~\ref{thm:main},  let $\alpha_{\text{sp}}(\widehat{\Delta})\le\frac{1}{c_0\PsiM}\sqrt{\frac{|\Omega|}{n\log n}}$. For large enough $c_0$, given any constant $\beta>0$, there exists constant $K_\beta>0$ such that, %assuming A\ref{ass:first}--\ref{ass:last}, if $|\Omega|=\boldsymbol{\Omega}(\PsiMsq n\log{n})$, 
under the assumptions in Theorem~\ref{thm:main}, w.p. $>1-4e^{-(1+\beta)\Psi^4_{\min}\log^2{n}}$:
\begin{equation*}%\Omega|}\sum_{(i,j)\in\Omega}B_G(\widehat{\Theta}_{ij},\Theta^*_{ij})\ge \kappamin\left(k_1\|\widehat{\Delta}\|^2_F-k_2{\mathcal{R}(\widehat{\Delta})}\frac{n\log n}{|\Omega|}\right)
\frac{mn}{|\Omega|}\sum_{{ij\in\Omega}}B_G(\widehat{\Theta}_{ij},\Theta^*_{ij}) \ge \kappamin\|\widehat{\Delta}\|_F^2,
\end{equation*}
where $\kappamin=e^{-\frac{2\eta\alpha^*}{\sqrt{mn}}}\left(K_\beta-\frac{64}{c_0}\sqrt{\frac{|\Omega|\kappa^2_{\mathcal{R}}(n,|\Omega|)}{n\log{n}}}\right)$.

\normalfont
As noted earlier, such an RSC result for the special case of squared loss under low--rank constraints was shown in \citet{negahban2012restricted}. We prove this theorem in Section~\ref{app:thm1}.  %However, proving the RSC condition for our general setting required a different and more involved proof technique.  

\label{thm:thm1}
\end{theorem}

\noindent \textbf{Remaining steps of the proof of Theorem \ref{thm:main}:} 
Thus, if  $\alpha_{\text{sp}}(\widehat{\Delta})\le\frac{1}{c_0\PsiM}\sqrt{\frac{|\Omega|}{n\log n}}$, and $\kappamin>0$, from Theorem~\ref{thm:thm1} and Lemma~\ref{lem:BFOmega}, 
%If $k_2\mathcal{R}(\widehat{\Delta})\frac{n\log n}{|\Omega|}>\frac{k_1}{2}\|\widehat{\Delta}\|_F^2$, then from Proposition~\ref{lem:inNu}, $\mathcal{R}(\widehat{\Delta})=\mathcal{R}(\mathcal{P}_{\mathcal{\overline{M}}^\perp}(\widehat{\Delta}))+\mathcal{R}(\mathcal{P}_\mathcal{\overline{M}}(\widehat{\Delta}))\le4\mathcal{R}(\mathcal{P}_\mathcal{\overline{M}}(\widehat{\Delta}))$
%Thus, we have the following:
%\begin{equation}
%\|\widehat{\Delta}\|_F^2<\frac{2k_2}{k_1}\mathcal{R}(\widehat{\Delta})\frac{n\log n}{|\Omega|}\le 4c^\prime\Psi(\mathcal{\overline{M}}) {\frac{n\log{n}}{|\Omega|}}\|\widehat{\Delta}\|_F
%\label{eq:case2}
%\end{equation}
%If on the other hand, if $k_2\mathcal{R}(\widehat{\Delta})\frac{n\log n}{|\Omega|}\le\frac{k_1}{2}\|\widehat{\Delta}\|_F^2$, then from Theorem~\ref{thm:thm1}, and Lemma~\ref{lem:BFOmega}, 
w.h.p.:
\small
\begin{align}
{\kappamin}\|\widehat{\Delta}\|^2_F\le\frac{mn}{|\Omega|}\sum_{{ij\in\Omega}}B_G(\widehat{\Theta}_{ij},\Theta^*_{ij})\le \frac{ 3\lambda\Psi(\mathcal{\overline{M}})}{2}\|\widehat{\Delta}\|_F
\label{eq:case1}
\end{align}

\normalsize
From~\eqref{eq:case3}~and~\eqref{eq:case1}, under assumptions of Theorem~\ref{thm:main},  w.p.  $>1-4e^{-(1+\beta)\Psi^4_{\min}\log^2{n}}$, we have:
\[\|\widehat{\Delta}\|^2_F\le \Psi^2(\mathcal{\overline{M}})\max\left\{\frac{3\lambda^2}{2\kappamin^2},\frac{\alpha^{*2}c_0^2n\log n}{|\Omega|}\right\}.
\]
%%%%%%%%%%%%%%%%%%%%%%%%%%%%%%
\subsection{Proof of Corollary~\ref{cor:corollary}}
From the definition of $\overline{\mathcal{M}}^\perp$ in~\eqref{eq:M}, we have $\overline{\mathcal{M}}=\text{span}\{\mathbf{u_i}x^\dagger,y\mathbf{v_j}^\dagger:x\in\mathbb{R}^n,\;y\in\mathbb{R}^m\}$. Let $P_{U^*}\in\mathbb{R}^{m\times m}$ and $P_{V^*}\in\mathbb{R}^{n\times n}$, be the projection matrices onto the column and row spaces ($U^*$, $V^*$) of $\Theta^*$, respectively.  We have, $\forall X\in\mathbb{R}^{m\times n}$, $X_{\overline{\mathcal{M}}}=P_{U^*}X+XP_{V^*}-P_{U^*}XP_{V^*}$. Also, $\text{rk}(P_{U^*})=\text{rk}(P_{V^*})=\text{rk}(\Theta^*)=r$. 
Thus, $\forall \Phi\in\overline{\mathcal{M}}$, $\text{rk}(\Phi)\le 2r$; and hence, \[\Psi(\overline{\mathcal{M}})=\underset{\Phi\in\overline{\mathcal{M}}\setminus\{0\}}{\text{sup}}\frac{\|\Phi\|_*}{\|\Phi\|_F}\le \sqrt{2r}.\text{ Further, }  \Psi_{\min}= 1.\]

Next, we use the following proposition by~\citet{negahban2012restricted}, to bound $\kappa_{\mathcal{R}}(n,|\Omega|)$ in Theorem~\ref{thm:main}.
\begin{lemma} If $\Omega\subset[m]\times[n]$ is sampled  using  uniform sampling and $|\Omega|>n\log{n}$, then for a Rademacher sequence $\{\epsilon_{ij},\forall (i,j)\in\Omega\}$, 
\[\mathbb{E}\Big[\frac{1}{|\Omega|}\|\sum_{ij\in\Omega}\sqrt{mn}\epsilon_{ij}e_ie_j^*\|_2\Big]\le 10\sqrt{\frac{n\log{n}}{|\Omega|}}.
\]
\normalfont This follows from Lemma~$6$~of~\citet{negahban2012restricted}, using $|\Omega|>n\log{n}$.\mybox
\end{lemma}
Thus, for large enough $c_0>640$, using $\kappa_{\mathcal{R}}(n,|\Omega|)=10\sqrt{\frac{n\log{n}}{|\Omega|}}$ in Theorem~\ref{thm:thm1}, for  some $K_\beta^\prime>0$ we have: 
\begin{equation}\kappamin=e^{-\frac{2\eta\alpha^*}{\sqrt{mn}}}\left(K_\beta-\frac{640}{c_0}\right)= K_\beta^\prime e^{-\frac{2\eta\alpha^*}{\sqrt{mn}}}.
\label{eq:temp3}
\end{equation}
Finally, to prove the corollary, we derive a bound on $\|\mathcal{P}_\Omega(X-g(\Theta^*))\|_2$ using the Ahlswede--Winter Matrix bound~(Appendix \ref{app:awmb}). Let $\phi(x)=\psi_2(x)=e^{x^2}-1$; and let $Y^{(ij)}\triangleq\sqrt{mn}(X_{ij}-g(\Theta^*_{ij}))e_ie_j^\dagger$, such that,
$\frac{\sqrt{mn}}{|\Omega|}\|\mathcal{P}_\Omega(X-g(\Theta^*))\|_2=\|\frac{1}{|\Omega|}\underset{ij\in\Omega}{\sum}Y^{(ij)}\|_2$.

From the equivalence of sub-Gaussian definitions in Lemma $5.5$ of~\citet{vershynin2010introduction}, there exists a constant $c_1$ such that $\|X_{ij}-g(\Theta^*_{ij})\|_\phi\le c_1 b$, $\forall ij$.  Since, $Y^{(ij)}$ has a single element, $\sqrt{mn}(X_{ij}-g(\Theta^*_{ij}))$, we have, $\|Y^{(ij)}\|_{\psi_2}\le c_1\sqrt{mn}b$. Further, 
\begin{align}
\nonumber&\mathbb{E}[Y^{(ij)^T}Y^{(ij)}]=\mathbb{E}[mn(X_{ij}-g(\Theta^*_{ij}))^2e_je^{*}_j]\overset{(a)}{=}mn \mathbb{E}_{(ij\in\Omega)}[\mathbb{E}_X[(X_{ij}-g(\Theta^*_{ij}))^2]e_je_j^*]\\
&\overset{(b)}{\le}mnb^2{\mathbb{E}_{(ij\in\Omega)}[e_je_j^*]}\overset{(c)}{=}mnb^2\frac{1}{n}I_{n\times n},
\label{eq:sigma}
\end{align}
where $(a)$ follows from Fubini's Theorem, $(b)$ follows as $(X_{ij}-g(\Theta^*_{ij}))$ is $b$-sub--Gaussian, and $(c)$ follows from the uniform sampling model. Similarly, $\mathbb{E}[Y^{(ij)}Y^{(ij)^T}]=mnb^2I_{m\times m}$. Define $\sigma_{ij}^2:=\max\{\mathbb{E}[Y^{(ij)^T}Y^{(ij)}],\mathbb{E}[Y^{(ij)}{Y^{(ij)}}^T]\}\le nb^2$

In Lemma~\ref{lem:awi}, using $\sigma^2:=\sum_{ij\in\Omega}\sigma_{ij}^2=n|\Omega|b^2$, $M=c_1\sqrt{mn}b\le c_1nb$, and $t=|\Omega|\delta$, we have:
\begin{equation*}
P\Big(\big\|\frac{1}{|\Omega|}\sum_{ij\in\Omega}Y^{(ij)}\big\|_2\ge \delta\Big)\le n^2\max\Big\{e^{-\frac{\delta^2|\Omega|}{4nb^2}},e^{-\frac{\delta|\Omega|}{2c_1nb}}\Big\}.
\end{equation*}
If $|\Omega|>cn\log{n}$ for large enough $c>0$, then for any constant  $C$, using $\delta=Cb\sqrt{\frac{n\log{n}}{|\Omega|}}$, 
\begin{align}
%&P\left(\frac{\sqrt{mn}}{|\Omega|}\|\mathcal{P}_\Omega(X-g(\Theta^*))\|_2\ge c_1b\sqrt{\frac{n\log{n}}{|\Omega|}}\right)\\&\nonumber\;\le e^{2\log{n}}\max\left\{e^{-{c_1^\prime}\log{n}},e^{-c_1^{\prime\prime}\sqrt{\frac{|\Omega|\log{n}}{n}}}\right\}\le e^{-C_2^\prime\log{n}}.
P\left(\frac{\sqrt{mn}}{|\Omega|}\|\mathcal{P}_\Omega(X-g(\Theta^*))\|_2\ge Cb\sqrt{\frac{n\log{n}}{|\Omega|}}\right)\le n^2e^{-\frac{C^2}{4}\log{n}}.
\end{align}

\noindent  Re-parameterizing the constants, we have for $\beta>0$, $\exists C_\beta>0$ such that w.p. $>1-e^{(1+\beta)\log{n}}$, $\frac{\sqrt{mn}}{|\Omega|}\|\mathcal{P}_\Omega(X-g(\Theta^*))\|_2\le C_\beta b\sqrt{\frac{n\log{n}}{|\Omega|}}$. Thus, using $\Psi_{\min}\ge1$, $\kappamin=K_\beta^\prime e^{-\frac{2\eta\alpha^*}{\sqrt{mn}}}$ (from~\eqref{eq:temp3}), and $\frac{\lambda}{2}:=C_\beta\sqrt{mn}b\sqrt{\frac{n\log{n}}{|\Omega|}}$ in Theorem~\ref{thm:main} leads to the corollary.

%%%%%%%%%%%%%%%%%%%%%%%%%%%

\begin{figure*}[t]
\centering
\begin{minipage}[htb]{0.31\textwidth}
\includegraphics[width=\columnwidth]{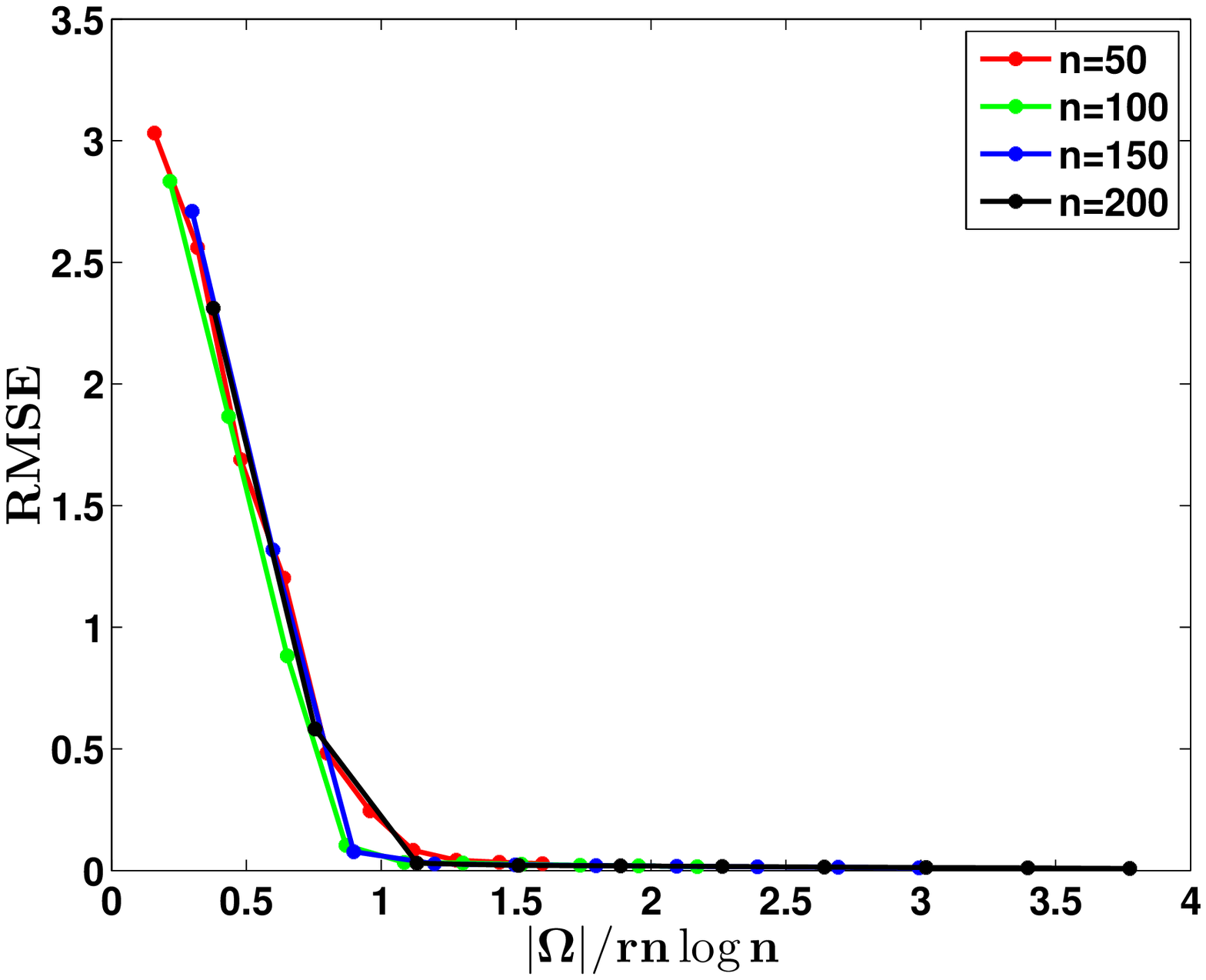} 
\end{minipage}
\begin{minipage}[htb]{0.31\textwidth}
\includegraphics[width=\columnwidth]{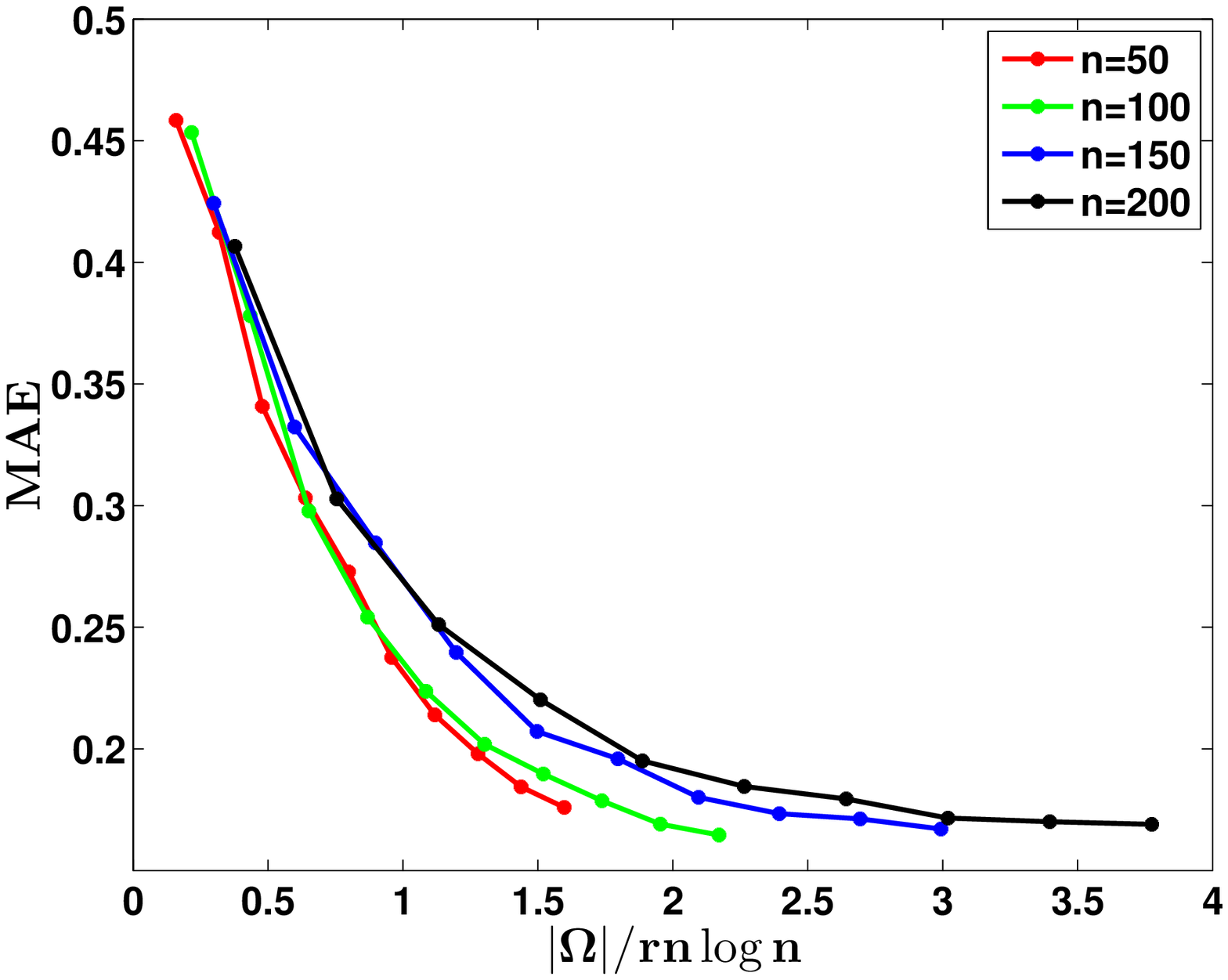} 
\end{minipage}
\begin{minipage}[htb]{0.31\textwidth}
\includegraphics[width=\columnwidth]{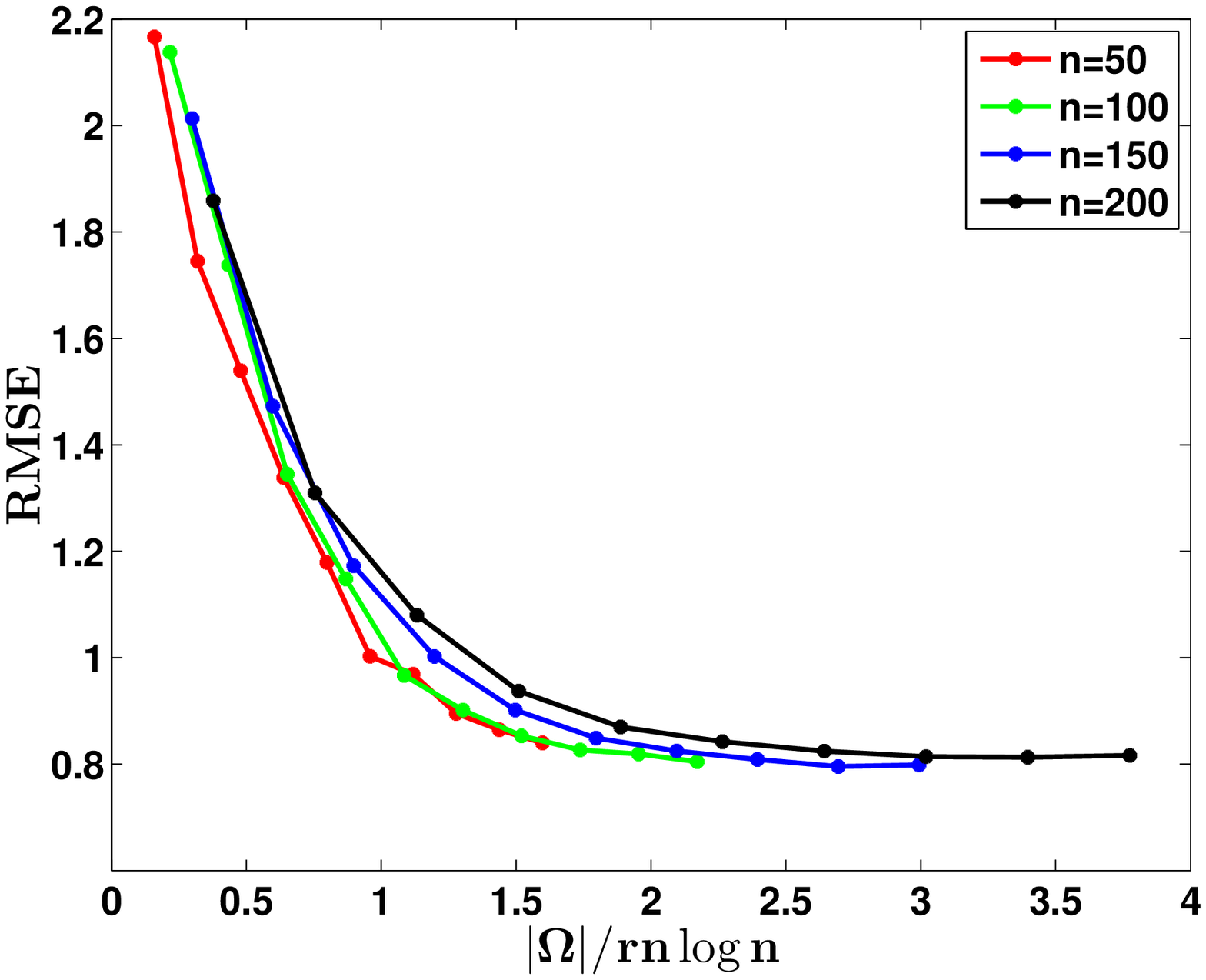} 
\end{minipage}
\caption{Appropriate error metric between observation matrix $X$, and the MLE estimate from \eqref{eq:main} $\widehat{X}$, plotted against ``normalized" sample size, when $X$ is generated from $(a)$ Gaussian, $(b)$ Bernoulli, and $(c)$ binomial distributions}
\end{figure*}

%%%%%%%%%%%%%%%%%%%%%%%%%%%

\subsection{Proof of Theorem \ref{thm:thm1}}\label{app:thm1}
This proof uses symmetrization arguments and contractions (\citet{ledoux1991probability} Ch.4\& 6). 
We observe that, $\forall\;(i,j)\in\Omega$, $\exists v_{ij}\in[0,1]$, s.t. \small
\begin{align}
\nonumber &B_G(\widehat{\Theta}_{ij},\Theta^*_{ij})=G(\widehat{\Theta}_{ij})-G(\Theta^*_{ij})-g(\Theta^*_{ij})(\widehat{\Theta}_{ij}-\Theta^*_{ij})\\&=\nabla^2G((1-v_{ij})\Theta^*_{ij}+v_{ij}\widehat{\Theta}_{ij})\widehat{\Delta}_{ij}^2\overset{(a)}{\ge}e^{-\frac{2\eta\alpha^*}{\sqrt{mn}}}\widehat{\Delta}_{ij}^2.
\label{eq:temp1}
\end{align}
\normalsize where $(a)$ holds as $|(1-v_{ij})\Theta^*_{ij}+v_{ij}\widehat{\Theta}_{ij}|\le\|\Theta^*\|_{\max}+\|\widehat{\Theta}\|_{\max}\le\frac{2\alpha^*}{\sqrt{mn}}$, and  $\nabla^2G(u)\ge e^{-\eta|u|}$~(Assumption \ref{ass:two}).

\begin{lemma}
Under Theorem~\ref{thm:thm1}, consider the subset 
\small \[\mathcal{E}=\Big\{\Delta\in\Nu: \alpha_{\text{sp}}(\Delta)\le\frac{1}{c_0\PsiM}\sqrt{\frac{|\Omega|}{n\log n}}, \|\Delta\|_F=1\Big\}.\]
\normalsize Given any constant $\beta>0$, there exists a constant $k_\beta>0$, such that w.p. $>1-4e^{-(1+\beta)\Psi^4_{\min}\log^2{n}}$, $\forall\;\Delta\in\mathcal{E}$:
\small
\begin{equation*}
\begin{split}
\Big|\frac{mn}{|\Omega|}\sum_{ij\in\Omega}\Delta_{ij}^2-1\Big|\le&\frac{16\mathcal{R}(\Delta)}{c_0\PsiM}\sqrt{\frac{|\Omega|\kappa^2_{\mathcal{R}}(n,|\Omega|)}{n\log{n}}}+\frac{k_\beta\mathcal{R}(\Delta)}{c_0^2\Psi(\overline{\mathcal{M}})}.
\end{split}
\end{equation*}
\normalsize \normalfont The proof is provided in Appendix \ref{app:temp2}. \mybox

\label{lem:temp2}
\end{lemma}
From the assumptions in Theorem~\ref{thm:thm1} and Proposition~\ref{lem:inNu},  $\frac{\widehat{\Delta}}{\|\widehat{\Delta}\|_F}\in\mathcal{E}$. Further, as $\widehat{\Delta}\in\Nu$, $\mathcal{R}(\widehat{\Delta})\le\mathcal{R}(\widehat{\Delta}_{\overline{\mathcal{M}}})
+\mathcal{R}(\widehat{\Delta}_{\overline{\mathcal{M}}^\perp})\le 4\mathcal{R}(\widehat{\Delta}_{\overline{\mathcal{M}}})\le 4\Psi(\overline{\mathcal{M}})\|\widehat{\Delta}\|_F$. 
%Using this in Lemma~\ref{lem:temp2}, we have  w.h.p.: 
%\begin{align}
%\nonumber 
%\frac{mn}{|\Omega|}&\sum_{ij\in\Omega}B_G(\widehat{\Theta}_{ij},\Theta^*_{ij})\ge 
%e^{-\frac{2\eta\alpha^*}{\sqrt{mn}}}\frac{mn}{|\Omega|}\sum_{ij\in\Omega}\widehat{\Delta}_{ij}^2\\
%\nonumber&\quad\quad\quad\overset{(a)}{\ge} e^{-\frac{2\eta\alpha^*}{\sqrt{mn}}}\|\widehat{\Delta}\|_F^2\Bigg(1-k_1\sqrt{\frac{\Psi^2(\overline{\mathcal{M}})n\log{n}}{|\Omega|}}\Bigg)\\&\quad\quad\quad\quad-e^{-\frac{2\eta\alpha^*}{\sqrt{mn}}}\|\widehat{\Delta}\|_F^2\frac{64}{c_0}\sqrt{\frac{|\Omega|\kappa^2_{\mathcal{R}}(n,|\Omega|)}{n\log{n}}}
%\label{eq:eq2}
%\end{align}
%where $(a)$ holds from Lemma~\ref{lem:temp2}, . 
%%Substituting this in Equation~\ref{eq:eq2}:
%%\begin{align}
%%\frac{mn}{|\Omega|}\sum_{ij\in\Omega}B_G(\widehat{\Theta}_{ij},\Theta^*_{ij})\ge e^{-\frac{2\eta\alpha^*}{\sqrt{mn}}}\|\widehat{\Delta}\|_F^2\left(1-\frac{640}{c_0}-k_1\sqrt{\frac{\Psi^2(\overline{\mathcal{M}})n\log{n}}{|\Omega|}}\right)
%%\end{align}
%
%Choosing $|\Omega|=c\Psi^2(\overline{\mathcal{M}})n\log{n}$, for large enough $c$, we can have $K_1:=1-k_1\sqrt{\frac{\Psi^2(\overline{\mathcal{M}})n\log{n}}{|\Omega|}}>0$, 
%\[\begin{split}&\frac{mn}{|\Omega|}\sum_{ij\in\Omega}B_G(\widehat{\Theta}_{ij},\Theta^*_{ij})\ge \kappamin \|\widehat{\Delta}\|_F^2\end{split}
%\]
%where $\kappamin=e^{-\frac{2\eta\alpha^*}{\sqrt{mn}}}\Big(K_1-\frac{64}{c_0}\sqrt{\frac{|\Omega|\kappa^2_{\mathcal{R}}(n,|\Omega|)}{n\log{n}}}\Big)$.
Using Lemma~\ref{lem:temp2}, and \eqref{eq:temp1}, and choosing $|\Omega|>c_0\Psi^2(\overline{\mathcal{M}})n\log{n}$, for large enough $c_0$, we have $K_\beta:=1-\frac{4k_\beta}{c_0^2}>0$. Finally,  using $\kappamin:=e^{-\frac{2\eta\alpha^*}{\sqrt{mn}}}\Big(K_\beta-\frac{64}{c_0}\sqrt{\frac{|\Omega|\kappa^2_{\mathcal{R}}(n,|\Omega|)}{n\log{n}}}\Big)$; if $\mu_\mathcal{L}>0$, then w.h.p.,
\begin{align}
\frac{mn}{|\Omega|}&\sum_{ij\in\Omega}B_G(\widehat{\Theta}_{ij},\Theta^*_{ij})\ge 
\kappamin \|\widehat{\Delta}\|_F^2.
\label{eq:eq2}
\end{align}

%% file: experiments.tex
\section{Experiments}\label{sec:experiments} 
We provide simulated experiments to corroborate our theoretical guarantees, focusing specifically on Corollary~\ref{cor:corollary}, where we consider the special case where the underlying parameter matrix is low--rank, but the underlying noise model for the matrix elements could be any of the general class of exponential family distributions. We look at three well known members of exponential family suitable for  different data--types, namely Gaussian, Bernoulli, and binomial, which are popular choices for modeling continuous valued, binary, and count valued data, respectively. 

\subsection{Experimental Setup}
We create low--rank ground truth parameter matrices, $\Theta^*\in\mathbb{R}^{m\times n}$ of sizes $n \in \{50,\;100,\;150,\;200\}$ (for simplicity we considered square matrices, $m=n$); we set the rank to $r=2\log{n}$. The observation matrices, $X$, are then sampled from the different members of exponential family distributions parameterized by $\Theta^*$. For each $n$, we uniformly sample a subset $\Omega$ entries of the observation matrix $X$, and estimate $\widehat{\Theta}$ from~\eqref{eq:main}. %We repeat this for various values of $|\Omega|$.

\noindent \textbf{Evaluation:}\\
For each member of the exponential family of distributions considered, we can measure the performance of our $M$--estimator in parameter space as $\frac{\|\widehat{\Theta}-\Theta^*\|_F^2}{\|\Theta^*\|_F^2}$, or in observation space using an appropriate error metric $\text{err}(\widehat{X},X)$, where $\widehat{X}$ is the maximum likelihood estimate of the recovered distribution, $\widehat{X}=\text{argmax}_X P(X|\widehat{\Theta})$ (we use RMSE, MAE in our plots). From our corollary, we require $|\Omega|=\mathcal{O}(rn\log{n})$ samples for consistent recovery, so we plot the error metric against the the ``normalized'' sample size, $\frac{|\Omega|}{rn\log{n}}$. For reasons of space, we only provide results for the error metric in observations space plotted against the the ``normalized'' sample size. The remainder of the results are provided in Appendix~$B$.  It can be seen from the plots that the error decays with increasing sample size, corroborating our consistency results; indeed $|\Omega|>1.5rn\log{n}$ samples suffice for the errors to decay to a very small value. Further, the aligning of the curves (for different $n$)  given the  ``normalized'' sample size corroborates the convergence rates as well. % Similar trends were observed for the norm of the parameters as well as shown in the Appendix~$B$. 

%\begin{figure}
%\centering
%\begin{minipage}[htb]{0.49\columnwidth}
%\includegraphics[width=\columnwidth]{Figures/RMSE_gauss_50_10_noscale.eps} 
%\end{minipage}
%\begin{minipage}[htb]{0.49\columnwidth}
%\includegraphics[width=\columnwidth]{Figures/RMSE_gauss_50_10_scale.eps} 
%\end{minipage}
%\end{figure}
%
%\begin{figure}
%\centering
%\begin{minipage}[htb]{0.49\columnwidth}
%\includegraphics[width=\columnwidth]{Figures/MAE1_bern_50_10_noscale.eps} 
%\end{minipage}
%\begin{minipage}[htb]{0.49\columnwidth}
%\includegraphics[width=\columnwidth]{Figures/MAE1_bern_50_10_scale.eps} 
%\end{minipage}
%\end{figure}
%
%\begin{figure}
%\centering
%\begin{minipage}[htb]{0.49\columnwidth}
%\includegraphics[width=\columnwidth]{Figures/rMSE_binom_50_10_noscale.eps} 
%\end{minipage}
%\begin{minipage}[htb]{0.49\columnwidth}
%\includegraphics[width=\columnwidth]{Figures/rMSE_binom_50_10_scale.eps} 
%\end{minipage}
%\end{figure}

%\begin{figure*}[H}
%\centering
%\begin{minipage}[htb]{0.32\textwidth}
%\includegraphics[width=\columnwidth]{Figures/RMSE_gauss_50_10_noscale.eps} 
%\end{minipage}
%\begin{minipage}[htb]{0.32\textwidth}
%\includegraphics[width=\columnwidth]{Figures/MAE1_bern_50_10_noscale.eps} 
%\end{minipage}
%\begin{minipage}[htb]{0.32\textwidth}
%\includegraphics[width=\columnwidth]{Figures/rMSE_binom_50_10_noscale.eps} 
%\end{minipage}
%\end{figure*}

%% file: appendix.tex
\appendix
\section{Proofs of Lemma}
\subsection{Proof of Lemma~\ref{lem:temp2}} \label{app:temp2}
Recall that $\Nu=\{\Delta:\mathcal{R}(\Delta_{\overline{\mathcal{M}}^\perp})\le 3\mathcal{R}(\Delta_{\overline{\mathcal{M}}})\}$. To prove Lemma~$4$, consider the nuclear norm ball $S_\mathcal{R}(t)=\{\Delta:\mathcal{R}(\Delta)\le t\}$.
\begin{enumerate}
\item Show that, $P\Bigg(\underset{\Delta\in\mathcal{E}\cap S_\mathcal{R}(t)}{\text{sup}}\left|\frac{mn}{|\Omega|}\sum_{ij\in\Omega}\Delta_{ij}^2-1\right|>\frac{8t}{c_0\PsiM}\sqrt{\frac{|\Omega|\kappa^2(n,|\Omega|)}{n\log{n}}}+\frac{k_\beta t}{2c_0^2\PsiM}\Bigg)$ is small; where $\kappa(n,|\Omega|)$ is a quantity that depends only on the dimensions $n$ and $|\Omega|$. This is done by:
\begin{enumerate}
\item Bounding the expectation, $\mathbb{E}\Big[\underset{\Delta\in\mathcal{E}\cap S_\mathcal{R}(t)}{\text{sup}}\left|\frac{mn}{|\Omega|}\sum_{ij\in\Omega}\Delta_{ij}^2-1\right|\Big]$
\item Showing an exponential decay of the tail.
\end{enumerate}
\item Then use a peeling argument~\cite{raskutti2010restricted} to derive at the result in Lemma~\ref{lem:temp2}.
\end{enumerate}

\subsubsection{Bounding Expectation}% ${\mathbb{E}\Big[\underset{\Delta\in\mathcal{E}\cap S_\mathcal{R}(t)}{\text{sup}}\left|\frac{mn}{|\Omega|}\sum_{ij\in\Omega}\Delta_{ij}^2-1\right|\Big]}$}
Note that $\forall\;\Delta\in\mathcal{E},\;\mathbb{E}[\frac{mn}{|\Omega|}\sum_{ij\in\Omega}\Delta_{ij}^2]=\|\Delta\|_F^2=1$. Thus, by using standard symmetrization argument~(Lemma~6.3~of~\cite{ledoux1991probability}, with a Rademacher sequence, $\{\epsilon_{ij},\forall\;ij\in\Omega\}$, we have:
\begin{equation}
\mathbb{E}\Big[\underset{\Delta\in\mathcal{E}\cap S_\mathcal{R}(t)}{\text{sup}}\Big|\frac{mn}{|\Omega|}\sum_{ij\in\Omega}\Delta_{ij}^2-1\Big|\Big]\le\frac{2mn}{|\Omega|}\mathbb{E}\Big[\underset{\Delta\in\mathcal{E}\cap S_\mathcal{R}(t)}{\text{sup}}\Big|\sum_{ij\in\Omega}\epsilon_{ij}\Delta_{ij}^2\Big|\Big]
\label{eq:e1}
\end{equation}
Also, $\forall\Delta\in\mathcal{E}$, $\phi_{ij}(\Delta)\triangleq\frac{\Delta_{ij}^2}{2\underset{\Delta\in\mathcal{E}}{\text{sup}}\|\Delta\|_{\max}}$ is a contraction, and $\forall\Delta\in\mathcal{E}$, $\|\Delta\|_{\max}=\frac{\alpha_{\text{sp}}(\Delta)}{\sqrt{mn}}\le\frac{1}{c_0\PsiM \sqrt{mn}}\sqrt{\frac{|\Omega|}{n\log{n}}}$. Thus, using Theorem~4.12 of~\cite{ledoux1991probability} in Equation~\ref{eq:e1}, we have:
\begin{align}
\begin{split}
\mathbb{E}\Big[\underset{\Delta\in\mathcal{E}\cap S_\mathcal{R}(t)}{\text{sup}}\Big|\frac{mn}{|\Omega|}\sum_{ij\in\Omega}\Delta_{ij}^2-1\Big|\Big]
&\le\frac{8}{c_0\PsiM}\sqrt{\frac{|\Omega|}{n\log{n}}}
\mathbb{E}\Bigg[\underset{\Delta\in\mathcal{E}\cap S_\mathcal{R}(t)}{\text{sup}}\Big|\frac{\sqrt{mn}}{|\Omega|}\Big\langle\sum_{ij\in\Omega}\epsilon_{ij}e_ie_j^*,\Delta\Big\rangle\Big|\Bigg]\\&
\overset{(a)}{\le}\frac{8t}{c_0\PsiM}\sqrt{\frac{|\Omega|}{n\log{n}}}
\mathbb{E}\Bigg[\frac{\sqrt{mn}}{|\Omega|}\mathcal{R}^*\Big(\sum_{ij\in\Omega}\epsilon_{ij}e_ie_j^*\Big)\Bigg]
\end{split}
\end{align}
where $(a)$ follows from Cauchy--Schwartz and as $\mathcal{R}(\Delta)\le t$. Note that $\mathcal{R}^*\Big(\sum_{ij\in\Omega}\epsilon_{ij}e_ie_j^*\Big)$ is independent of $\Delta$ and depends only on $n$ and $|\Omega|$. Let $\kappa(n,|\Omega|)\ge\mathbb{E}\Bigg[\frac{\sqrt{mn}}{|\Omega|}\mathcal{R}^*\Big(\sum_{ij\in\Omega}\epsilon_{ij}e_ie_j^*\Big)\Bigg]$ be a suitable upper bound. 
\begin{equation}
\mathbb{E}\Big[\underset{\Delta\in\mathcal{E}\cap S_\mathcal{R}(t)}{\text{sup}}\Big|\frac{mn}{|\Omega|}\sum_{ij\in\Omega}\Delta_{ij}^2-1\Big|\Big]{\le}\frac{8t}{c_0\PsiM}\sqrt{\frac{|\Omega|\kappa^2(n,|\Omega|)}{n\log{n}}}
\end{equation}
%\max\left\{1,\sqrt{\frac{n\log{n}}{|\Omega|}}\right\}\le k_1^\prime t, and $(b)$ a is consequence of the following proposition from~\cite{negahban2012restricted} and using the condition $|\Omega|=\boldsymbol{\Omega}(\Psi^2(\mathcal{\overline{M}})n\log{n})>n\log{n}$. 

\subsubsection{Tail Behavior}
Let $G_t(\Omega)\triangleq\underset{\Delta\in\mathcal{E}\cap S_\mathcal{R}(t)}{\text{sup}}\Big|\frac{mn}{|\Omega|}\sum_{ij\in\Omega}\Delta_{ij}^2-1\Big|$. Let $\Omega^\prime\subset [m]\times [n]$ be another set of indices that differ from $\Omega$ in exactly one element. We then have:
\begin{align*}
\begin{split}
&G_t(\Omega)-G_t(\Omega^\prime)=\underset{\Delta\in\mathcal{E}\cap S_\mathcal{R}(t)}{\text{sup}}\Big|\frac{mn}{|\Omega|}\sum_{ij\in\Omega}\Delta_{ij}^2-1\Big|-\underset{\Delta\in\mathcal{E}\cap S_\mathcal{R}(t)}{\text{sup}}\Big|\frac{mn}{|\Omega|}\sum_{kl\in\Omega^\prime}\Delta_{kl}^2-1\Big|\\
&\le\frac{mn}{|\Omega|}\underset{\Delta\in\mathcal{E}\cap S_\mathcal{R}(t)}{\text{sup}}\bigg(\Big|\sum_{ij\in\Omega}\Delta_{ij}^2-1\Big|-\Big|\sum_{kl\in\Omega^\prime}\Delta_{kl}^2-1\Big|\bigg)\le\frac{mn}{|\Omega|}\underset{\Delta\in\mathcal{E}\cap S_\mathcal{R}(t)}{\text{sup}}\bigg(\Big|\sum_{ij\in\Omega}\Delta_{ij}^2-\sum_{kl\in\Omega^\prime}\Delta_{kl}^2\Big|\bigg)\\
&\le\frac{2mn}{|\Omega|}\underset{\Delta\in\mathcal{E}\cap S_\mathcal{R}(t)}{\text{sup}}\|\Delta\|_{\max}^2\le\frac{2}{c_0^2\Psi^2(\mathcal{\overline{M}})n\log{n}}
\end{split}
\end{align*}
By similar arguments on $G_t(\Omega^\prime)-G_t(\Omega)$, we conclude that $|G_t(\Omega)-G_t(\Omega^\prime)|\le \frac{2}{c_0^2\Psi^2(\mathcal{\overline{M}})n\log{n}}$.
Therefore, using Mc Diarmid's inequality, we have $P(|G_t(\Omega)-\mathbb{E}[G_t(\Omega)]|>\delta)\le 2\exp\left(-\frac{c_0^4\delta^2\Psi^4(\mathcal{\overline{M}})n^2\log^2{n}}{2|\Omega|}\right)$. Fix $\delta=\frac{2 k_1 t}{c_0^2\PsiM}$ for appropriate constant $k_1$. Recall that $\Psi_{\min}=\underset{X \setminus\{0\}}{\inf}{\frac{\mathcal{R}(X)}{\|X\|_F}}\le\PsiM$. Using $|\Omega|\le n^2$,
\[P\Big(G_t(\Omega)>\frac{8t}{c_0\PsiM}\sqrt{\frac{|\Omega|\kappa^2(n,|\Omega|)}{n\log{n}}}+\frac{2 k_1 t}{c_0^2\PsiM}\Big)\le 2\exp\left({-2k_1^2 t^2\Psi^2_\text{min}\log^2{n}}\right)
\]
%Further, as $|\Omega|=\mathcal{O}(\Psi^2(\mathcal{\overline{M}})n\log{n})$
%\[P\left(G_t(\Omega)>\frac{8t}{c_0\PsiM}\sqrt{\frac{|\Omega|\kappa^2(n,|\Omega|)}{n\log{n}}}+\frac{k_1}{2}t\sqrt{\frac{n\log{n}}{|\Omega|}}\right)\le 2\exp(-{c_1^{\prime\prime} t^2n\log{n}})\]

\subsubsection{Peeling Argument}
Consider the following sets,  $S_\ell=\{\Delta\in\mathcal{E}:2^{\ell-1}\Psi_{\min}\le\mathcal{R}(\Delta)\le 2^\ell\Psi_{\min}\}$, for all (integers) $\ell\ge 1$. Since, $\forall\Delta\in\mathcal{E}$, $\mathcal{R}(\Delta)\ge\Psi_{\min}\|\Delta\|_F=\Psi_{\min}$, for each $\Delta\in\mathcal{E}$, $\Delta\in S_\ell$ for some $\ell\ge1$. 
Further, if for some $\Delta\in\mathcal{E}$, $\Big|\frac{mn}{|\Omega|}\sum_{ij\in\Omega}\Delta_{ij}^2-1\Big|>\frac{16\mathcal{R}(\Delta)}{c_0\PsiM}\sqrt{\frac{|\Omega|\kappa^2(n,|\Omega|)}{n\log{n}}}+\frac{4 k_1 \mathcal{R}(\Delta)}{c_0^2\PsiM}$, then for some $\ell$:
\begin{align}
\begin{split}
\Big|\frac{mn}{|\Omega|}\sum_{ij\in\Omega}\Delta_{ij}^2-1\Big|&>\frac{16( 2^{\ell-1})\Psi_{\min}}{c_0\PsiM}\sqrt{\frac{|\Omega|\kappa^2(n,|\Omega|)}{n\log{n}}}+\frac{4 k_1 2^{\ell-1}\Psi_{\min}}{c_0^2\PsiM}\\&=\frac{8(2^{\ell}\Psi_{\min})}{c_0\PsiM}\sqrt{\frac{|\Omega|\kappa^2(n,|\Omega|)}{n\log{n}}}+\frac{2 k_1( 2^{\ell}\Psi_{\min})}{c_0^2\PsiM}
\end{split}
\end{align}
Thus, 
\begin{align}
\begin{split}
&P\bigg(\underset{\Delta\in\mathcal{E}}{\text{sup}}\bigg|\frac{mn}{|\Omega|}\sum_{ij\in\Omega}\Delta_{ij}^2-1\bigg|>\frac{16\mathcal{R}(\Delta)}{c_0\PsiM}\sqrt{\frac{|\Omega|\kappa^2(n,|\Omega|)}{n\log{n}}}+\frac{4 k_1 \mathcal{R}(\Delta)}{c_0^2\PsiM}\bigg)\\
&\quad\le\sum_{\ell=1}^\infty P\bigg(G_{2^\ell}(\Omega)>\frac{8(2^{\ell}\Psi_{\min})}{c_0\PsiM}\sqrt{\frac{|\Omega|\kappa^2(n,|\Omega|)}{n\log{n}}}+\frac{2 k_1( 2^{\ell}\Psi_{\min})}{c_0^2\PsiM}\bigg)
\le\sum_{\ell=1}^\infty 2\exp{(-2k_1^2 2^{2l}\Psi^4_\text{min}\log^2{n})}\\&\quad\overset{(a)}{\le}\sum_{\ell=1}^\infty 2\exp(-{4\log{2}\;k_1^2\ell \Psi^4_{\min}\log^2{n}})\le \frac{2e^{-4k_1^2 \Psi^4_{\min}\log^2{n}}}{1-e^{-4k_1^2 \Psi^4_{\min}\log^2{n}}}\le 4e^{-4k_1^2 \Psi^4_{\min}\log^2{n}}
\end{split}
\end{align}
where $(a)$ follows as $x\ge\log{x}$ for $x>1$, and the last step holds for $n>1$. The lemma follows by re-parametrization of constants in terms of $\beta$.

\subsection{Proof of Lemma \ref{lem:BFOmega}}\label{app:BFOmega}
Let $\widehat{\Delta}=\widehat{\Theta}-\Theta^*$. %To avoid clutter, we use the notation $X_{\mathcal{\overline{M}}}=\mathcal{P}_{\mathcal{\overline{M}}}(X)$, and $X_{\mathcal{\overline{M}}^\perp}=\mathcal{P}_{\mathcal{\overline{M}}^\perp}(X)$.
\begin{align}
\mathcal{R}(\widehat{\Theta})=\mathcal{R}\big({\Theta}^*+\widehat{\Delta}_\mathcal
{\overline{M}}+\widehat{\Delta}_{\mathcal{\overline{M}}^\perp}\big)\ge
\mathcal{R}\big({\Theta}^*+\widehat{\Delta}_{\mathcal{\overline{M}}^\perp}\big)-
\mathcal{R}\big(\widehat{\Delta}_\mathcal{\overline{M}}\big)=\mathcal{R}\big({\Theta}^*\big)
+\mathcal{R}\big(\widehat{\Delta}_{\mathcal{\overline{M}}^\perp}\big)-
\mathcal{R}\big(\widehat{\Delta}_\mathcal{\overline{M}}\big)
\label{eq:temp}
\end{align}
The above inequalities hold due to triangle inequality, and  decomposability of $\mathcal{R}$ over $\Theta^*\in\mathcal{M}$ and $\Delta_{\mathcal{\overline{M}}^\perp}\in\mathcal{\overline{M}}^\perp$.
\begin{align}
\nonumber\frac{mn}{|\Omega|}\sum_{(i,j)\in\Omega}B_G(\widehat{\Theta}_{ij},\Theta^*_{ij})
&=\frac{mn}{|\Omega|}\big[\!\sum_{(i,j)\in\Omega}G(\widehat{\Theta}_{ij})-X_{ij}\widehat{\Theta}_{ij}-G(\Theta_{ij}^*)+X_{ij}\Theta^*_{ij}+\langle \mathcal{P}_\Omega(X-g(\Theta^*),\widehat{\Delta}\rangle\big]\\
\nonumber&\overset{(a)}{\le}\lambda\mathcal{R}(\Theta^*)-\lambda\mathcal{R}(\widehat{\Theta})+\frac{mn}{|\Omega|}\mathcal{R}^*(\mathcal{P}_\Omega(X-g(\Theta^*)))\mathcal{R}(\widehat{\Delta})\\
\nonumber&\overset{(b)}{\le}\lambda\mathcal{R}(\widehat{\Delta}_\mathcal{\overline{M}})-
\lambda\mathcal{R}(\widehat{\Delta}_{\mathcal{\overline{M}}^\perp})+\frac{\lambda}{2}\mathcal{R}(\widehat{\Delta}_\mathcal{\overline{M}}+\widehat{\Delta}_{\mathcal{\overline{M}}^\perp})
\overset{(c)}{\le}\frac{3\lambda}{2}\mathcal{R}(\widehat{\Delta}_\mathcal{\overline{M}})-
\frac{\lambda}{2}\mathcal{R}(\widehat{\Delta}_{\mathcal{\overline{M}}^\perp})\\
&\le\frac{3\lambda\Psi(\mathcal{\overline{M}})}{2}\|\Theta^*-\widehat{\Theta}_\mathcal{\overline{M}}\|_F\le\frac{3\lambda\Psi(\mathcal{\overline{M}})}{2}\|\Theta^*-\widehat{\Theta}\|_F
\label{eq:eqn1}
\end{align}
where $(a)$ follows as $\widehat{\Theta}$ is the minimizer of~\eqref{eq:main} and using Cauchy Schwartz, $(b)$ follows from 0\eqref{eq:temp} and using $\frac{mn}{|\Omega|}\mathcal{R}^*(\mathcal{P}_\Omega(X-g(\Theta^*))\le\frac{\lambda}{2}$, and $(c)$ follows from triangle inequality.$\hfill\Box$

\subsection{Ahlswede--Winter Matrix Bound (Extension)}\label{app:awmb}
The Orlicz norm of a random matrix $Z\in\mathbb{R}^{m\times n}$ w.r.t to a convex, differentiable and monotonically increasing function, $\phi(x):\mathbb{R}^+\to\mathbb{R}$ as follows:
\begin{align*}
\|Z\|_{\phi}\triangleq &\text{inf}\{t\ge0: \mathbb{E}\left[\phi\left({|\langle Z,Z^\prime\rangle|}/{t})\right)\right]\le 1 ,\\&\quad\forall\;Z^\prime\in\mathbb{R}^{m\times n},\; and\;Z^\prime_{ij}\in[0,1]\}
\end{align*}
\begin{lemma} [Ahlswede-Winter Matrix Bound]
Let $Z^{(1)},Z^{(2)},\ldots,Z^{(K)}$ be random matrices of dimensions $m\times n$. Let $\|Z^{(i)}\|_\phi\le M$, $\forall i$. Further, $\sigma_i^2=\max\{\|\mathbb{E}[Z^{(i)^T}Z^{(i)}]\|_2,\|\mathbb{E}[Z^{(i)}Z^{(i)^T}]\|_2\}$, and $\sigma^2=\sum_{i=1}^K\sigma_i^2$, then:
\[P\left(\|\sum_{i=1}^KZ^{(i)}\|_2\ge t\right)\le mn\max\left\{e^{-\frac{t^2}{4\sigma^2}},e^{-\frac{t}{2M}}\right\}
\]
\label{lem:awi}
\end{lemma}
The above lemma is an extension noted by~\cite{vershynin2009note} (Theorem~$1$ and a later remark) for the matrix bounds resulting from~\cite{ahlswede2002strong}. 

\section{Additional Experimental Results}\label{app:exp}
We provide the additional experimental results where we compare the error of the estimate in the parameter space. We plot the results first against the proportion of the total entries sampled, $\frac{|\Omega|}{mn}$ (Figure on left), and then against the ``normalized" sample size, $\frac{|\Omega|}{rn\log{n}}$ (Figures on right). We observe trends similar to those observed in Section~$5$. Again, we find that the curves (for different $n$) given the ``normalized" sample size, align and converge (left), corroborating the theoretical results. Note that, the curves do not align when plotted against, unnormalized sample size (right). Further, as with errors in observation space, with $|\Omega|> 1.5rn\log{n}$ samples, the errors parameter space also decay to a sufficiently small value.

\begin{figure}[htb]
\centering
\begin{minipage}[htb]{0.49\textwidth}
\includegraphics[width=\columnwidth]{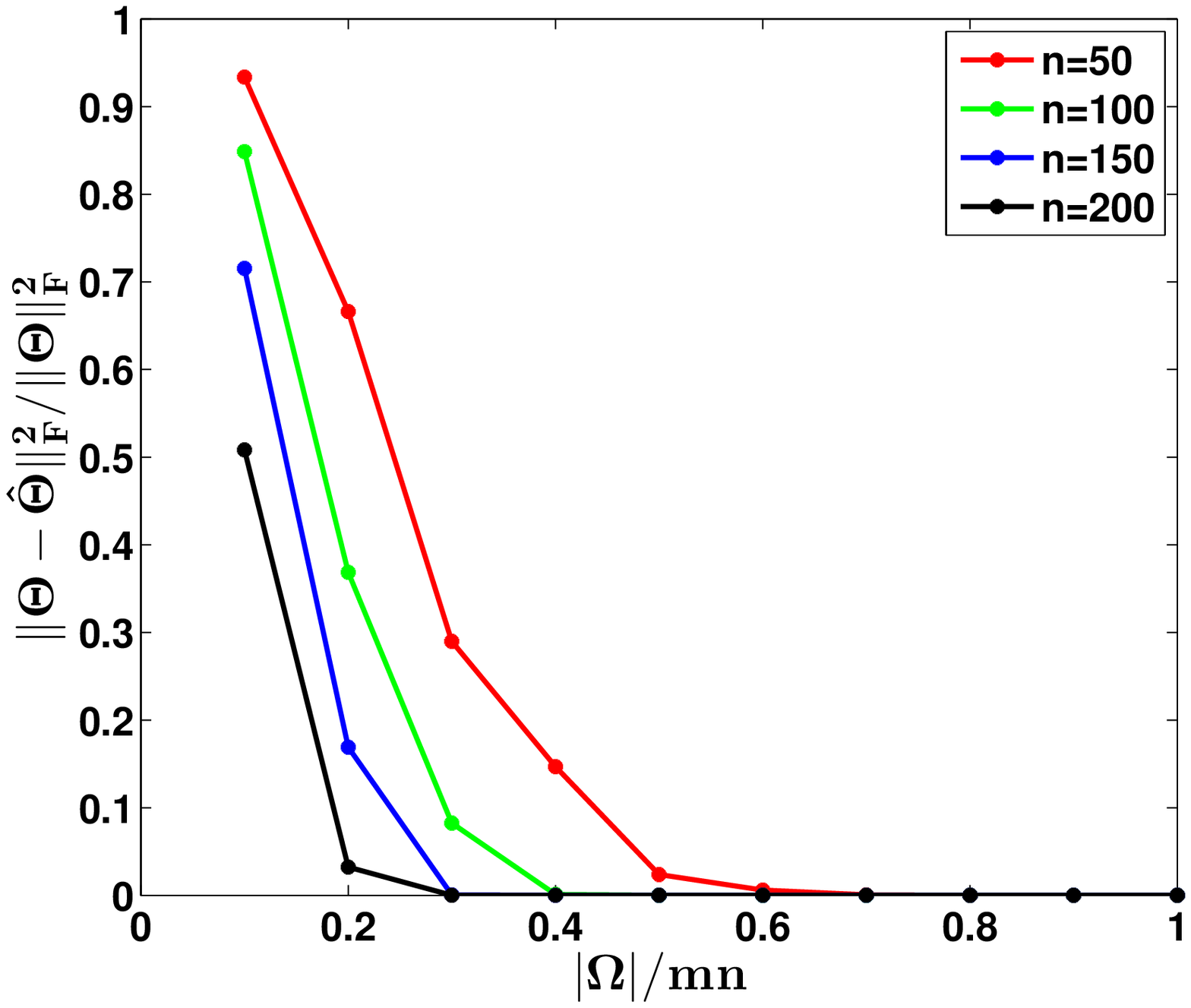} 
\end{minipage}
~
\begin{minipage}[htb]{0.49\textwidth}
\includegraphics[width=\columnwidth]
{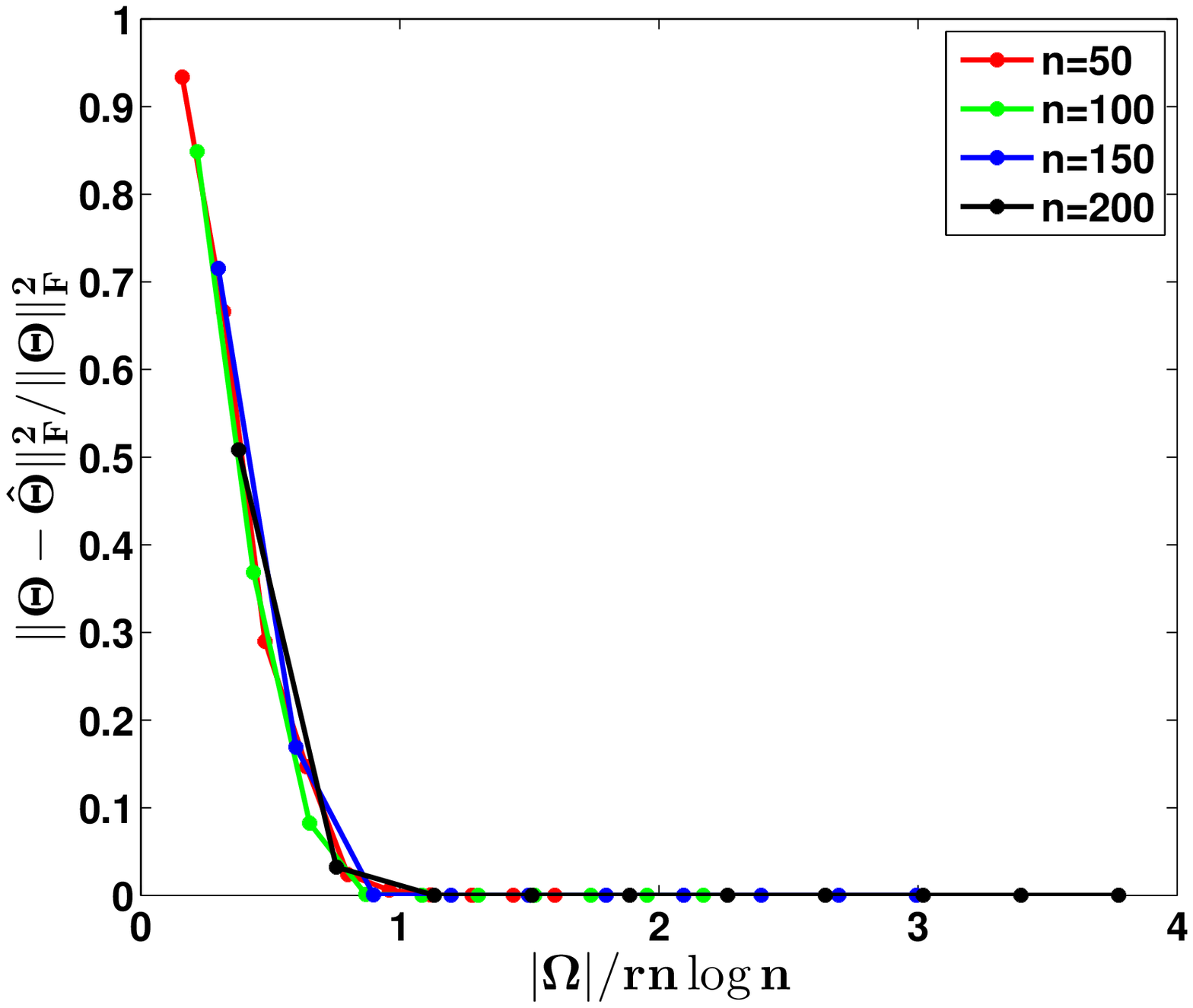} 
\end{minipage}
\caption{Parameter Error when measured  $(a)$ against proportion of the sampled values, and $(b)$ against the `normalized" sample size, when the distribution of the observations $P(X|\Theta^*)$, is Gaussian}
\end{figure}

\begin{figure}[H]
\centering
\begin{minipage}[htb]{0.49\textwidth}
\includegraphics[width=\columnwidth]{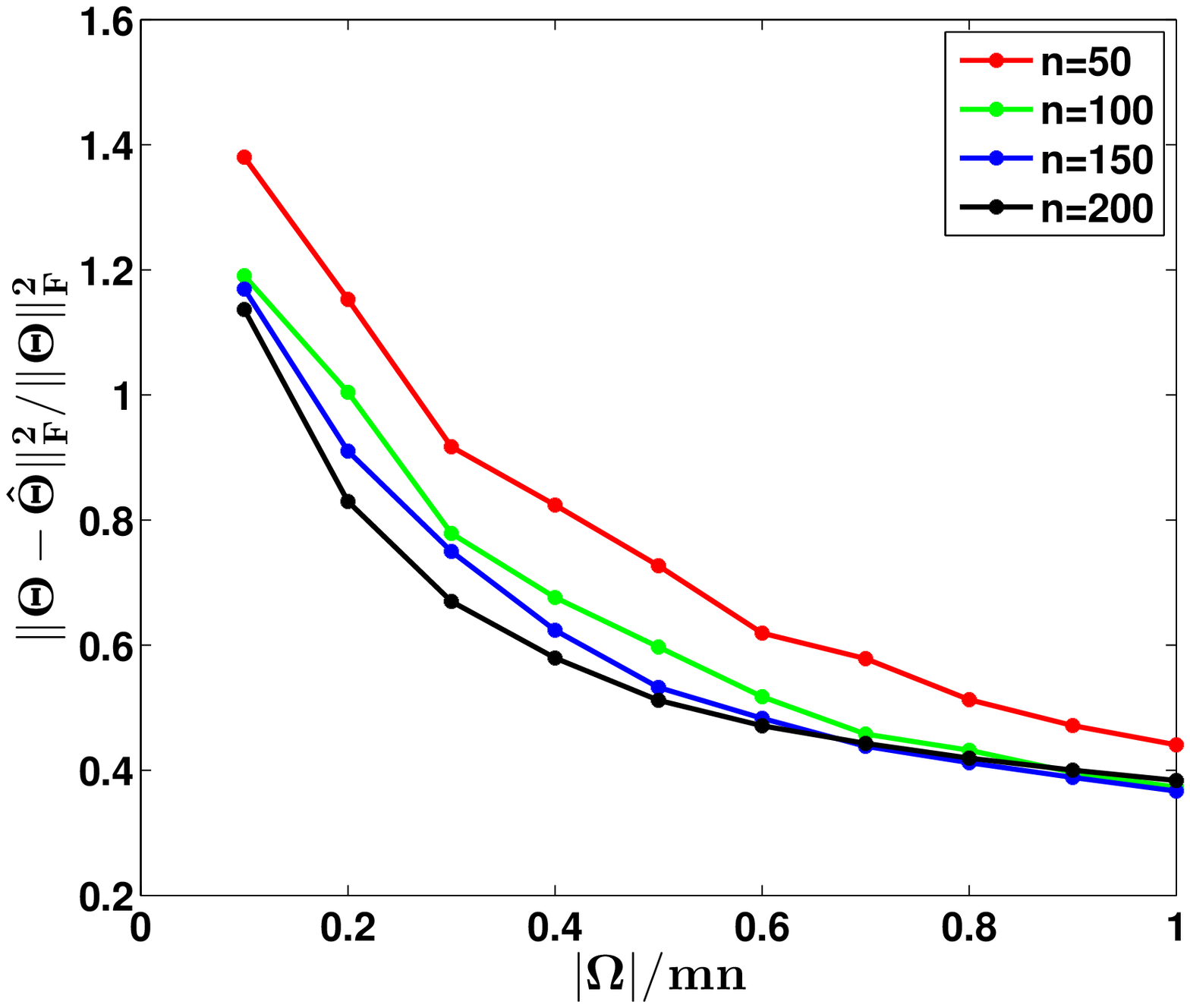} 
\end{minipage}
~
\begin{minipage}[htb]{0.49\textwidth}
\includegraphics[width=\columnwidth]
{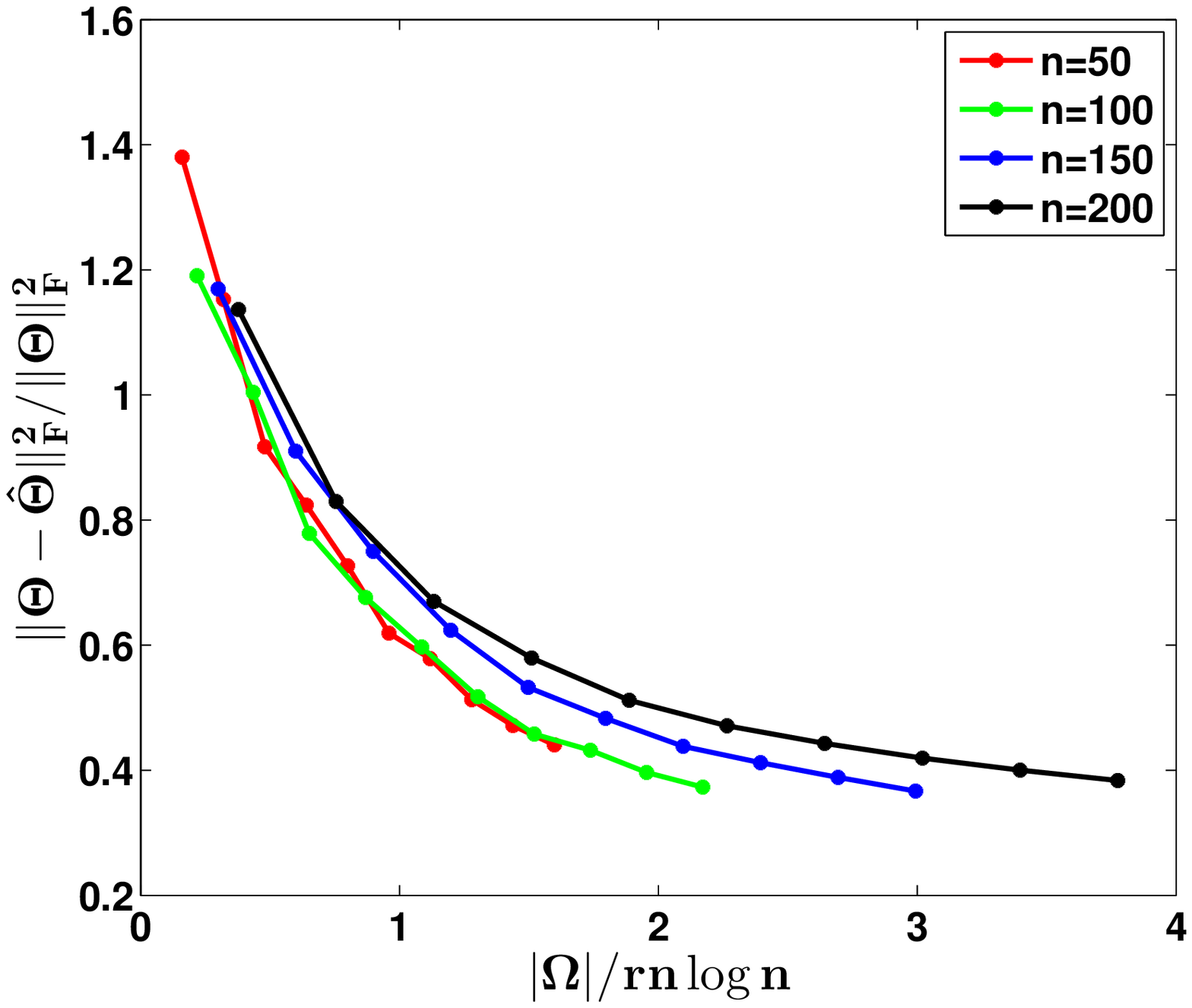} 
\end{minipage}
\caption{Parameter Error when measured  $(a)$ against proportion of the sampled values, and $(b)$ against the `normalized" sample size, when the distribution of the observations $P(X|\Theta^*)$, is Bernoulli}
\end{figure}

\begin{figure}[H]
\centering
\begin{minipage}[htb]{0.49\textwidth}
\includegraphics[width=\columnwidth]{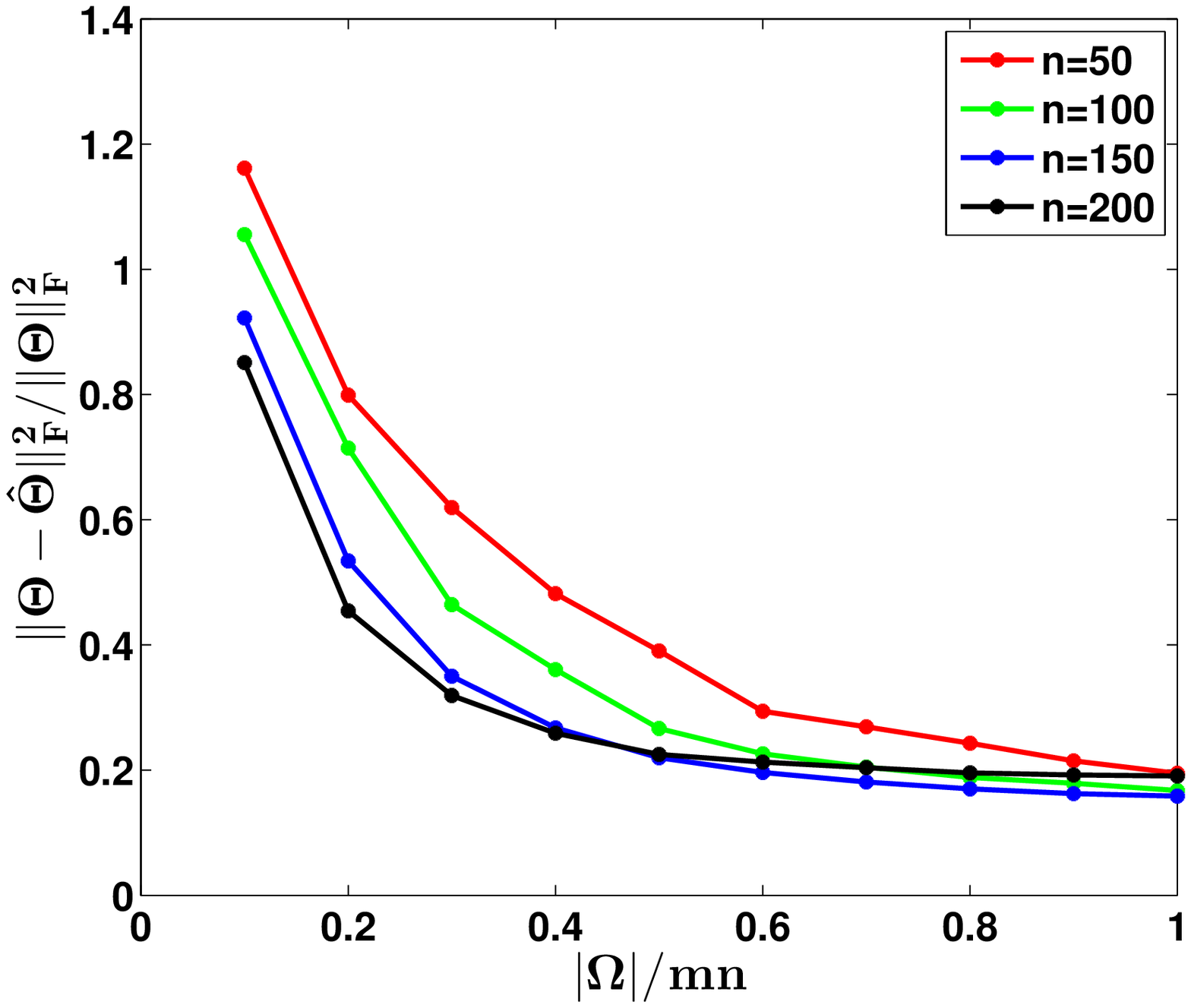} 
\end{minipage}
~
\begin{minipage}[htb]{0.49\textwidth}
\includegraphics[width=\columnwidth]
{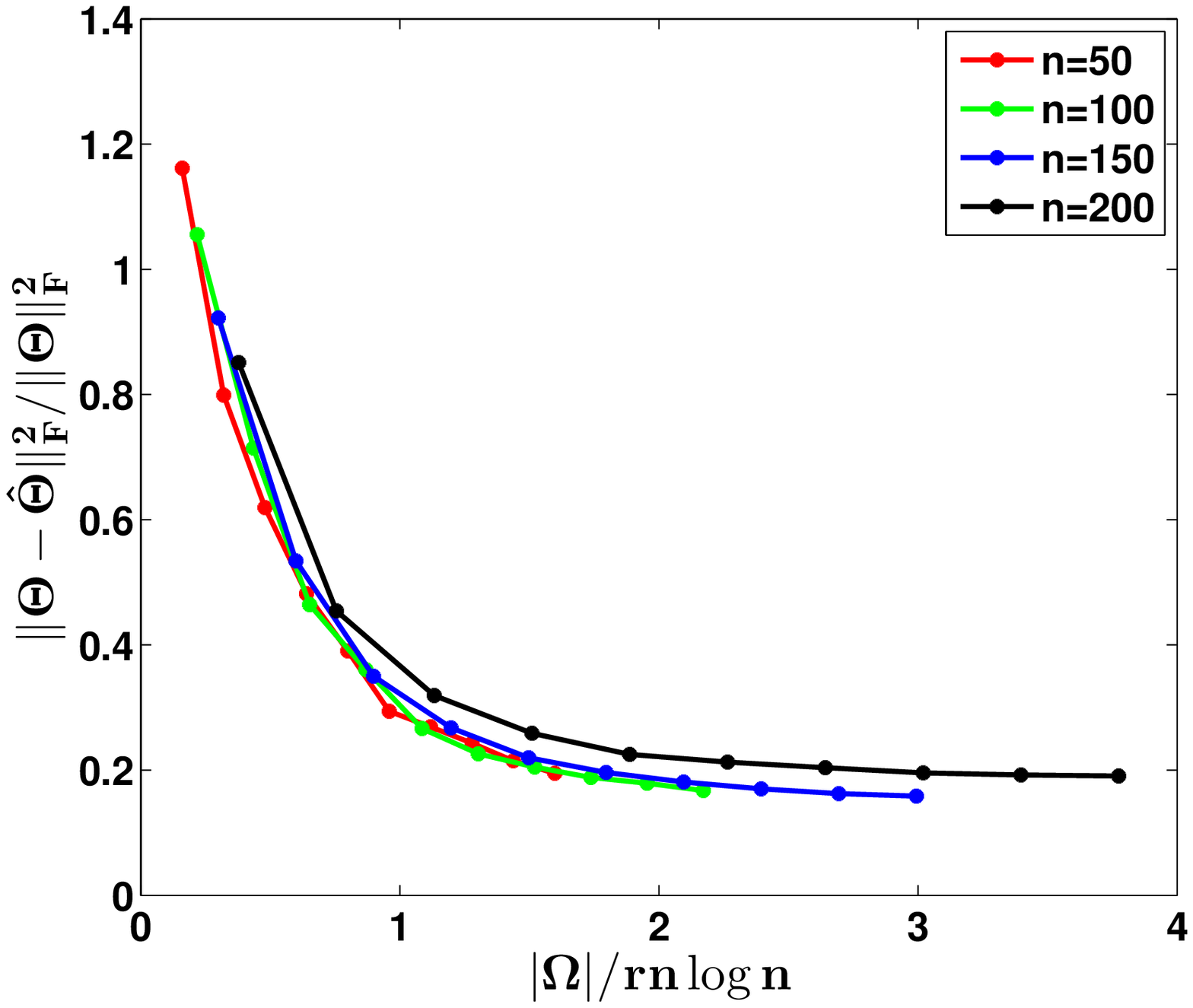} 
\end{minipage}
\caption{Parameter Error when measured  $(a)$ against proportion of the sampled values, and $(b)$ against the `normalized" sample size, when the distribution of the observations $P(X|\Theta^*)$, is Binomial}
\end{figure}